\documentclass[11pt]{article}

\usepackage[final]{acl}

\usepackage{times}
\usepackage{latexsym}
\usepackage{float}
\usepackage{placeins}
\usepackage{booktabs}
\usepackage{amsmath}
\usepackage{amssymb}
\usepackage[T1]{fontenc}
\usepackage[utf8]{inputenc}
\usepackage{verbatim}
\usepackage{microtype}
\usepackage{csquotes}
\usepackage{inconsolata}

\usepackage{graphicx}

\title{Language Defaults Before Content: \\ A Causal Layer Map of Multilingual Generation}
\author{
Shenbin Qian \and Yves Scherrer \\
Department of Informatics \\
University of Oslo \\
\texttt{\{shenbinq, yves.scherrer\}@ifi.uio.no}
}

\begin{document}
\maketitle
\begin{abstract}
Multilingual generation requires large language models (LLMs) to determine both what to say and in what language to say it, yet these two commitments are difficult to disentangle from model outputs alone. We use translation as a controlled probe: for multi-parallel segments that a model translates well in one language direction but poorly in another, content is held fixed while language varies. Using activation patching, we write residual-stream states from successful runs into failing ones and compare content-matched states with content-independent controls, allowing us to trace content-specific and language-related signals across layers. Across three LLMs of different scales and architectures, we find a consistent three-stage pattern of multilingual generation. Intermediate layers first express a default English signal; content information becomes causally available one layer later; and the final target language is resolved only at later layers. The ordering of the English default before content also appears in a reading-comprehension task. Our sparse autoencoder analyses reveal a further dissociation: the intermediate language signal is only weakly represented by sparse features, whereas a later language distinction is clearly sparse, but resists intervention at the patched position. These results provide a mechanistic account of how language selection and content information emerge and interact during multilingual generation.
\end{abstract}

\section{Introduction}

Large language models (LLMs) are trained on heavily English-centric data~\citep{Touvron2023-ww,kew-etal-2024-turning,guo-etal-2025-large}, yet are routinely used for generation tasks across many languages~\citep{ZhangBiao2023,wang-etal-2025-multilingual,lai-etal-2024-llms}. This creates a basic mechanistic puzzle: when generating text, a multilingual model must determine both \textit{what to say} (content) and \textit{in what language to say it} (language). These decisions are entangled in the final output, however, so behavioral evaluations alone cannot reveal where in the network they arise, or in which order they are resolved.

We make these two commitments experimentally visible by exploiting multi-parallel data, where the same content exists in many languages. Specifically, we construct a \emph{contrastive instance} from one parallel segment translated in two directions: one direction that the model translates well (\textbf{HIGH}) and the other direction that it translates poorly (\textbf{LOW}). The two directions may differ in source, target, or both, but always express the same content, so language varies while content is fixed. Using activation patching~\citep{Heimersheim2024-gs}, we copy the residual-stream state at the last-token position of the input prompt from HIGH into LOW, one layer at a time, and ask what the donor state supplies. When the two targets differ, the patch can switch the LOW output into the HIGH target language. When they coincide, no language switch is possible, and any quality gain against the LOW reference reflects content.

However, a successful HIGH$\rightarrow$LOW patch need not reveal what kind of information is transferred. Even though HIGH and LOW are content-matched, the donor state may carry both content-specific information and content-independent signals, such as the intended output language. We therefore compare the \textit{matched} HIGH donor with three controls: a HIGH state from another instance (\textit{shuffled}), the average HIGH state (\textit{mean}), and the difference between average HIGH and LOW states (\textit{mean\_diff}), added as a steering vector. Their relative effects let us distinguish information that depends on the particular content from information shared across successful multilingual generations, and trace how these components emerge across depth.

Our findings reveal a three-stage account of multilingual generation in LLMs. First, the residual stream carries a default English signal at intermediate layers. Second, content information emerges one layer later, where the content-matched donor first separates from the controls. This content transfer is not specific to translation, as we observe the same phenomenon in a reading-comprehension task. Third, the language ultimately produced by the model resolves at later layers, with the amount of language-specific information transferred depending on the distance between the source and target languages. We replicate this three-stage pattern across three LLMs of different sizes and architectural designs, suggesting that the ordering of these stages is a general property of multilingual generation, while their precise locations and geometry vary across models.

We further use sparse autoencoders~\citep{huben2024sparse} to test whether these language commitments correspond to sparse features. The intermediate English-default transition behaves causally like a single linear direction, yet no sparse feature set reproduces its effect. The later distinction among related target languages shows the converse pattern: it is sparsely represented, but last-token interventions do not change the model’s language choice. Together, these results show that a language signal can be sparsely encoded without being causally effective at the same position, while a causally effective signal need not be cleanly represented by sparse features. Our main contributions are:

\begin{itemize}\setlength{\itemsep}{1pt}
\item A donor-control methodology ($\S$\ref{sec:setup}) for cross-lingual activation patching---\textit{matched}, \textit{shuffled}, \textit{mean}, and additive \textit{mean\_diff} donors---that decomposes patching effects into content-specific and content-independent components, applied to $22$ translation and $9$ reading-comprehension contrastive pairs.
\item A three-stage, layer-localised account ($\S$\ref{decomp_results}) of what a single last-token residual state carries, with converging evidence from patching onsets and the logit lens, replicated across three LLMs of different sizes and architectures.
\item A feature-level dissociation ($\S$\ref{sec:feature}) between sparse representability and last-token causality, together with evidence that autoencoder reconstruction fidelity bounds what this class of analysis can establish.
\end{itemize}

\section{Related Work}

Recent work on multilingual processing in LLMs suggests a layered picture: early layers map inputs into a shared representation space, middle layers operate within that space, and late layers project it into a specific output language~\citep{zeng-etal-2025-converging,ravisankar-etal-2026-map}. In English-centric models, this shared space itself appears biased toward English, as intermediate representations of non-English prompts can decode preferentially into English~\citep{wendler-etal-2024-llamas}. Two strands of work further develop complementary parts of this picture, corresponding closely to the two signals we aim to separate. The first studies mechanisms that determine the \emph{output language}. Prior work identifies language-specific neurons and subspaces~\citep{tang-etal-2024-language,kojima-etal-2024-multilingual}, and shows that manipulating them can steer which language a model generates~\citep{cao-etal-2024-exploring,raihan-chowdhury-2026-causal}. Failures of this control mechanism can produce fluent responses in an unintended language, a phenomenon commonly termed language confusion~\citep{marchisio-etal-2024-understanding,nie-etal-2025-mechanistic}. The second strand asks what multilingual representations preserve \textit{across} languages: shared semantic structure in middle layers~\citep{tezuka-inoue-2025-transfer}, language-agnostic knowledge~\citep{meng2022,jin-etal-2025-exploring,wang-etal-2025-lost-multilinguality,dumas-etal-2025-separating}, and content-independent task or function vectors that transfer instructions across prompts~\citep{todd2024function,hendel-etal-2023-context}.

Our work connects these two lines through a controlled success--failure contrast. Rather than identifying language- or content-related structure in isolation, we compare the same underlying content in two translation directions: one in which the model succeeds (HIGH), and one in which it fails (LOW). We then use donor-controlled activation patching to ask which signals are transferred when a successful residual-stream state is written into a failing computation, and to distinguish content-specific from content-independent effects. To our knowledge, prior work has not used such matched success--failure interventions to study how content information and language-related signals are jointly organized within the same residual state.

\section{Experimental Setup}
\label{sec:setup}

\paragraph{Data.} We use FLORES+~\citep{maillard-etal-2024-findings} for translation and Belebele~\citep{bandarkar-etal-2024-belebele} for reading comprehension. Both are multi-parallel: segment $i$ is the same content in every language, aligned by an identifier. Belebele's passages are drawn from FLORES sentences, so both tasks share the same material. When FLORES+ \texttt{devtest} ($1{,}012$ segments) yields too few contrastive instances, we top up from NTREX-128~\citep{federmann-etal-2022-ntrex} through an identical pipeline in $\S$\ref{sec:contrasts}.

\paragraph{Tasks and metrics.} For machine translation (MT), we adopt the prompt template, \verb|{src_lang}: {src}|$\backslash$\verb|n{tgt_lang}: {tgt}|, from \citet{ZhangBiao2023} in a 5-shot setting, with demonstrations separated by blank lines and the test item terminating at the trailing \verb|{tgt_lang}:|. We write $Q(\hat{y}, y)$ for the segment-level quality of a hypothesis $\hat{y}$ against a reference $y$, instantiated with two metrics: chrF++~\citep{popovic-2017-chrf} and xCOMET-XL~\citep{guerreiro-etal-2024-xcomet} (which additionally conditions on the source).\footnote{BLEU~\citep{papineni-etal-2002-bleu} is excluded as a corpus-level statistic, as our unit of analysis is the segment.} Belebele is evaluated zero-shot with a multiple-choice (MC) question answering template (\verb|P:|/\verb|Q:|/\verb|A:B:C:D:|/\verb|Answer:|), scored by the next-token log-probability of the four option letters, each verified to be a single token. Accuracy is computed based on whether the most probable next token is the correct option letter.

\paragraph{Contrastive instances.}
\label{sec:contrasts}
We formally define a contrastive instance as an item $i$ for which the model performs well in one language condition but poorly in another one, while the underlying content is held fixed. For MT, given an ordered pair of directions ($\textsc{low}, \textsc{high}$), we retain each segment that satisfies:
\begin{equation}\label{eq:selection}
q_{\textsc{low}}(i) < \tau_{\textsc{low}}
\;\wedge\;
q_{\textsc{high}}(i) \ge \tau_{\textsc{high}},
\end{equation}
where $q_{\textsc{low}}(i)$ and $q_{\textsc{high}}(i)$ are the xCOMET-XL scores $Q(\hat{y}, y)$ of the model's unpatched translation of segment $i$ in each direction. We set thresholds separately for each language pair (Appendix~\ref{sec:pairs}), requiring a margin of $\tau_{\textsc{high}}-\tau_{\textsc{low}}\ge 0.10$ and a certain amount ($\ge50$) of surviving instances. A single global cutoff would either eliminate LOW pairs or admit near-ties in HIGH ones. For Belebele, the criterion is binary: incorrect in the LOW language and correct in the HIGH language.

This yields $22$ MT contrastive pairs in three groups, defined by which parts of the translation directions are shared and therefore what a patch can transfer. \textbf{Same-target} pairs ($7$; e.g., \texttt{kas}$\rightarrow$\texttt{eng} vs.\ \texttt{hin}$\rightarrow$\texttt{eng}) share the target language, so no target-language change is possible and any improvement reflects content transfer. \textbf{Different-target, same-source} pairs ($9$; e.g., \texttt{eng}$\rightarrow$\texttt{nno} vs.\ \texttt{eng}$\rightarrow$\texttt{swe}) share the source, making content transfer uninformative and isolating the target-language decision. \textbf{Disjoint} pairs ($6$; e.g., \texttt{kas}$\rightarrow$\texttt{cmn} vs.\ \texttt{hin}$\rightarrow$\texttt{eng}) differ in both source and target, so donor controls are needed to separate content from language effects. Belebele contributes $9$ additional contrastive pairs over the same or closely related languages. Full pair definitions, counts, and thresholds are reported in  Tables~\ref{tab:contrasting-MC} and~\ref{tab:contrasting-MT} in Appendix~\ref{sec:pairs}.

\paragraph{Models.} Our primary model is \texttt{gemma-3-4b-pt}, a $34$-layer decoder-only base model with a $2560$-dimensional residual stream \citep{Gemma_Team2025-wv}. We use the pretrained rather than instruction-tuned model, so that the behaviour reflects pretraining-induced language competence rather than alignment. We replicate on \texttt{Qwen3.5-9B-Base}~\citep{qwen3.5} and \texttt{Llama-3.1-8B}~\citep{Grattafiori2024-ot} in \S\ref{generalization}. Inference details can be found in Appendix~\ref{sec:inf_details}.

\subsection{Activation Patching}
\label{sec:patching}

We write $\mathbf{h}^{(\ell)}(x) \in \mathbb{R}^{d}$ for the residual-stream output of block $\ell$ at the \emph{last token position} of prompt $x$, and abbreviate $\mathbf{h}_i^{(\ell)} = \mathbf{h}^{(\ell)}(x_i^{\textsc{high}})$ and $\tilde{\mathbf{h}}_i^{(\ell)} = \mathbf{h}^{(\ell)}(x_i^{\textsc{low}})$ for the two prompts of the contrastive instance $i$. $\bar{\mathbf{h}}^{(\ell)}_{\!\textsc{high}} = \frac{1}{N}\sum_{j} \mathbf{h}_j^{(\ell)}$ and $\bar{\mathbf{h}}^{(\ell)}_{\!\textsc{low}} = \frac{1}{N}\sum_{j} \tilde{\mathbf{h}}_j^{(\ell)}$ denote the average over the $N$ instances of a HIGH and a LOW pair.

\paragraph{Patch site.} We intervene on a single vector of one layer, at the last token position of the prompt only.\footnote{Patching the shared source span for same-source pairs produces no change in the reference log-probability, so only last-token patching is used in the rest of the paper.} Under both tasks, that position is a fixed template token: the trailing \verb|{tgt_lang}:| for MT and \verb|Answer:| for MC. This makes copying of a last-token vector from one prompt into another positionally meaningful. Each intervention therefore edits a single $d$-dimensional vector, $\mathbf{h}^{(\ell)}(x_i^{\textsc{low}})$, out of the $|x| \times L$ residual vectors of the LOW forward pass for a prompt of $|x|$ tokens and a model of $L$ layers. All other states are computed normally. %so any change in the output must originate from this one position and layer.

\paragraph{Intervention.} We use two types of patch at the patch site. A \emph{replacement} patch overwrites the LOW state with a \emph{donor vector} $\mathbf{v}_i^{(\ell)}$ obtained from HIGH forward passes (Eq.~\ref{eq:replace}). An \emph{additive} patch keeps the LOW state and shifts it along a \emph{steering direction} $\boldsymbol{\delta}^{(\ell)}$ (Eq.~\ref{eq:add}):
\begin{equation}
\label{eq:replace}
\mathbf{h}^{(\ell)}\!\left(x_i^{\textsc{low}}\right) \;\longleftarrow\; \mathbf{v}_i^{(\ell)} ,
\end{equation}
\begin{equation}
\label{eq:add}
\mathbf{h}^{(\ell)}\!\left(x_i^{\textsc{low}}\right) \;\longleftarrow\; \tilde{\mathbf{h}}_i^{(\ell)} + \alpha\,\boldsymbol{\delta}^{(\ell)} ,
\end{equation}
where $\alpha$ scales the step ($\alpha = 1$ unless stated otherwise). The donor conditions below differ only in how $\mathbf{v}_i^{(\ell)}$ or $\boldsymbol{\delta}^{(\ell)}$ is constructed. Each layer is patched independently. We never patch two at once, so a result curve over $\ell$ reports the marginal causal contribution of that layer's last-token state.

\paragraph{Donor conditions.} A gain under Eq.~\ref{eq:replace} is ambiguous, since the donor may have supplied the segment's content, or only a content-free signal about how to execute the task. We therefore compare four donors, of which only the first sees the test segment. \textbf{Matched} is the HIGH activation of the same instance, $\mathbf{v}_i^{(\ell)} = \mathbf{h}_i^{(\ell)}$ (the positive control, replicating a standard cross-lingual patch). \textbf{Shuffled} is the HIGH activation of a different instance, $\mathbf{v}_i^{(\ell)} = \mathbf{h}_{\pi(i)}^{(\ell)}$ with $\pi$ a seeded derangement. This ensures that no gain here comes from the content. \textbf{Mean} is the per-layer average of all $N$ HIGH activations, rescaled to the average per-instance norm,
\begin{equation}
\mathbf{v}^{(\ell)} = \gamma^{(\ell)} \bar{\mathbf{h}}^{(\ell)}_{\!\textsc{high}},
\quad
\gamma^{(\ell)} = \frac{\frac{1}{N}\sum_{j}\lVert \mathbf{h}_j^{(\ell)}\rVert}{\lVert \bar{\mathbf{h}}^{(\ell)}_{\!\textsc{high}}\rVert},
\end{equation}
which is the same for every instance $i$ and thus instance-independent by construction. $\gamma^{(\ell)}$ restores the norm lost through averaging, so a null effect cannot be attributed to reduced donor magnitude. \textbf{Mean\_diff} is the only additive condition (Eq.~\ref{eq:add}), with steering direction $\boldsymbol{\delta}^{(\ell)} = \bar{\mathbf{h}}^{(\ell)}_{\!\textsc{high}} - \bar{\mathbf{h}}^{(\ell)}_{\!\textsc{low}}$. Unlike the three replacement donors, it preserves the instance's own state, testing whether the effect relies on a single linear direction. Both $\pi$ and the means are computed over the full instance list from a fixed seed. We sweep $18$ of Gemma's $34$ layers, densely where the effects manifest and sparsely elsewhere, after an initial sweep of all layers for one pair. Detailed patching procedure is in Appendix~\ref{sec:patch_procedure}. 

\paragraph{Signal measure.} \label{sec:readouts}
We read three signals to measure a language flip or a content transfer: (i) \emph{Quality}: $\Delta_{\textsc{low}} = Q(\hat{y}^{\mathrm{p}}, y^{\textsc{low}}) - Q(\hat{y}^{\mathrm{c}}, y^{\textsc{low}})$ is used to compare the patched $\hat{y}^{\mathrm{p}}$ and clean $\hat{y}^{\mathrm{c}}$ outputs against the reference of the LOW direction $y^{\textsc{low}}$. For different-target pairs, the same quality difference $\Delta_{\textsc{high}}$ is used against the HIGH direction's gold translation $y^{\textsc{high}}$. We report both in chrF++ in the main text, with xCOMET-XL values given alongside and in full in Appendix~\ref{sec:add_results}. %When read jointly they separate a language switch (\textit{shuffled} and \textit{mean} rising on $\Delta_{\textsc{high}}$ give the switch-only floor) from content transfer (\textit{matched} above that floor).
(ii) \emph{Output language}: we first utilise fastText \texttt{lid.176}~\citep{joulin-etal-2017-bag} to detect the output language and compute the corpus-level fraction of generations in the LOW or HIGH target language at each layer and condition. We report it as other if neither LOW nor HIGH target is detected. We call it \emph{flip fraction}. Language identification may not be reliable on single short outputs in closely related languages, so we count target-exclusive characters, i.e., exclusive letters for same-script siblings (Norwegian \ae/\o\ against Swedish \"a/\"o) and Unicode script membership for different-script targets as $c_{\textsc{high}}$ or $c_{\textsc{low}}$. We mark an instance as flipped at the first layer ($\ell \ge 8$) where the HIGH-target exclusive share $c_{\textsc{high}}/(c_{\textsc{low}}{+}c_{\textsc{high}})$ reaches $0.5$. %, where its characters are half depleted relative to the clean output. 
We call this layer the flip \textit{onset layer}. The onset layer of a pair is the median of that layer over all flipped instances. Flip fraction and onset layer answer different questions: how far the flip goes, and where it begins. We report them jointly in $\S$\ref{non_eng_onset}. (iii) For Belebele, we rely on \textit{accuracy} difference and whether the patched prediction matches the \textit{HIGH pass's own prediction} of that instance to detect content transfer.

%We utilise the above readout signals to compare between donor conditions at a layer to capture language flip or content transfer. 
Since all conditions are evaluated on the same instances, we compute paired per-instance differences and average them across instances. We assess whether this mean difference is non-zero using a two-sided bootstrap test with $10{,}000$ resamples.% over instances. Resampling instances this way tests the intervention rather than the particular sample of sentences. Layers and conditions are never pooled.

\begin{figure*}[t]
  \centering
  \includegraphics[width=0.32\textwidth]{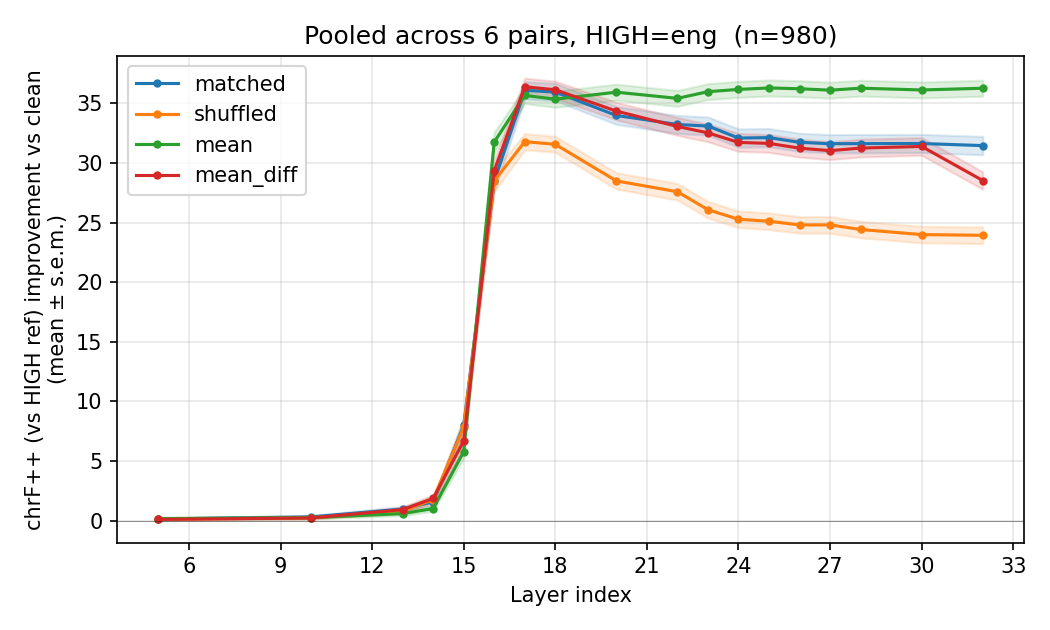}\hfill
  \includegraphics[width=0.32\textwidth]{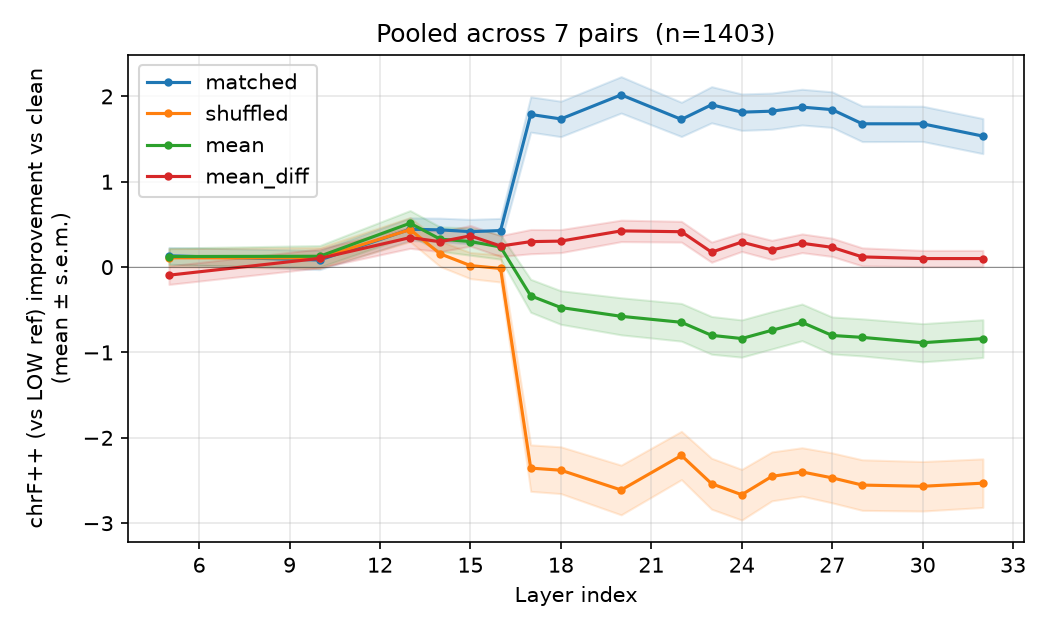}\hfill
  \includegraphics[width=0.32\textwidth]{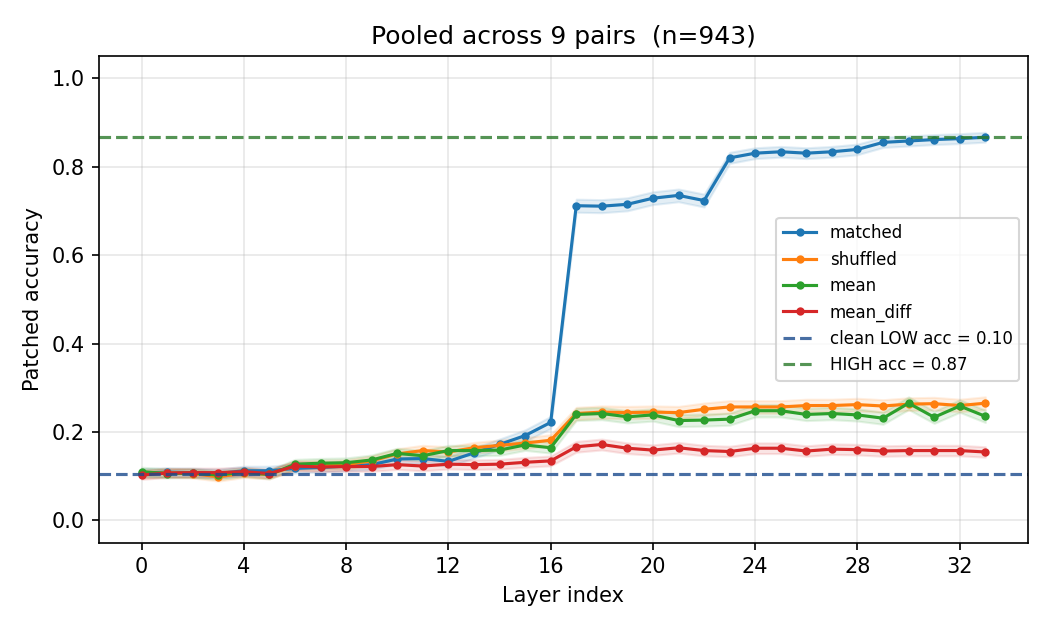}
  \caption{Per-layer effect of last-token patching in \texttt{gemma-3-4b-pt}, pooled over instances and read per donor condition. \textbf{Left}: $\Delta_{\textsc{high}}$ (chrF++ against the HIGH target's gold translation) for the six different-target pairs whose HIGH direction targets English ($n{=}980$). All four donors rise together at L15--16; only at L17 does \textit{matched} separate. \textbf{Middle}: $\Delta_{\textsc{low}}$ for the seven same-target pairs ($n{=}1403$), where no language flip is possible; the \textit{matched}--\textit{shuffled} gap opens at L17. \textbf{Right}: Belebele accuracy for nine language contrasts ($n{=}943$); the clean baseline is $0.10$ and the HIGH pass scores $0.87$. xCOMET-XL versions of the two MT panels are in Appendix~\ref{sec:add_results}.}
  \label{fig:decomp}
\end{figure*}

\subsection{Sparse Autoencoders} 

We employ sparse autoencoders (SAEs) to investigate the language commitments in \S\ref{sec:feature}. An SAE is trained to reconstruct a model activation $\mathbf{h} \in \mathbb{R}^{d}$ as a sparse, non-negative combination drawn from an overcomplete dictionary of $d_{\mathrm{SAE}} \gg d$ directions \citep{bricken2023monosemanticity,huben2024sparse}:
\begin{equation}
\label{eq:sae}
\begin{split}
\mathbf{a}(\mathbf{h}) &= \sigma\!\left(W_{\mathrm{enc}}\mathbf{h} + \mathbf{b}_{\mathrm{enc}}\right), \\
\hat{\mathbf{h}} &= \sum\nolimits_{f} a_f(\mathbf{h})\,\mathbf{d}_f + \mathbf{b}_{\mathrm{dec}} .
\end{split}
\end{equation}
Each dictionary element $\mathbf{d}_f \in \mathbb{R}^{d}$, a column of $W_{\mathrm{dec}}$, is a \emph{feature}, and the scalar $a_f(\mathbf{h}) \ge 0$ indicates how strongly that feature is active, on this activation. The motivating hypothesis is that features are more nearly monosemantic than neurons, so each responds to one interpretable condition, which can be checked by inspecting the inputs that activate it most \citep{templeton2024scaling}. For the SAEs we use, the JumpReLU of Gemma Scope 2 \citep{gemmascope2} would zero out any activation below a learned threshold (per-feature), and the TopK autoencoders for Qwen and Llama~\citep{Deng2026-yx,He2024-vk} keep only the $k$ largest features. Therefore, only the hundreds ($L_0$) out of the $10^4$--$10^5$ monosemantic features need to be checked for their responsibility. However, SAE reconstruction is never perfect. The residual $\mathbf{h} - \hat{\mathbf{h}}$ is invisible to any feature-level manipulation, so the relative $L_2$ fidelity $1 - \lVert\hat{\mathbf{h}}-\mathbf{h}\rVert/\lVert\mathbf{h}\rVert$ bounds what such an SAE-feature analysis can establish.

\section{Decomposition Results}
\label{decomp_results}

We report three effects separated by layer: a content-independent switch to English at L15-16, the onset of content transfer at L17, and the resolution of a specific non-English target at L24.

\paragraph{The English flip is content-independent.}
Figure~\ref{fig:decomp} (left) pools the six different-target pairs whose HIGH direction targets English. We can see before L15, patching is inert ($\Delta_{\textsc{high}} < 2$ chrF++ for every condition). Between L15 and L16 all four conditions jump together: $\Delta_{\textsc{high}}$ rises from $8.0$ to $28.5$ for \textit{matched}, $7.8$ to $28.5$ for \textit{shuffled}, $5.8$ to $31.8$ for \textit{mean} and $6.7$ to $29.4$ for \textit{mean\_diff}. xCOMET-XL moves in step ($0.06 \rightarrow 0.18$ for \textit{matched}, and $0.18$--$0.19$ for the other three). The paired difference \textit{matched}$-$\textit{shuffled} is $0.15$ chrF++ at L15 ($p = 0.81$) and $-0.02$ at L16 ($p = 0.96$), which suggests that whether the donor comes from a matched or a random segment makes no measurable difference to this effect. \textit{Shuffled}, \textit{mean} and \textit{mean\_diff}, all three reproduce the full effect, indicating that what the last-token residual supplies at L16 is not the test segment but a single, instance-independent, additive direction. The measurement of language output also agrees, as the median onset layer is L15--17 under every condition for all six pairs, and $68$--$94\%$ of instances end in English.

\paragraph{Content transfer begins at L17.} \label{sec:content_transfer}
Figure~\ref{fig:decomp} (middle) shows $\Delta_{\textsc{low}}$ for the same-target pairs. We see at L17, \textit{matched} rises to $+1.79$ chrF++ while \textit{shuffled} falls to $-2.35$ and \textit{mean} to $-0.34$. The paired \textit{matched}--\textit{shuffled} gap is $0.43$ chrF++ at L16 ($p = 0.04$) and $4.43$ at L17 ($p < 10^{-3}$), and it persists to L32. xCOMET-XL reproduces both the onset and the ordering ($0.05$ vs.\ $-0.01$ at L17). The absolute effect is small because the LOW output is already in the right language and only its content improves. However, the sign structure is unambiguous. The wrong-segment donor is actively harmful, which is what a content channel predicts.

Belebele makes the same point without any surface metric. \textit{Matched} moves accuracy from $0.22$ at L16 to $0.71$ at L17, while \textit{shuffled} and \textit{mean} plateau at $0.26$ and $0.24$, chance for a four-way choice (paired gap $0.47$, $p<10^{-3}$ at L17). The share of instances on which the patched LOW pass reproduces the \emph{HIGH pass's own prediction} rises from $0.30$ at L16 to $0.80$ at L17, against a flat $0.28$ for \textit{shuffled}. This suggests that \textit{matched} begins to transfer the answer at L17. The fact that both tasks with unrelated formulations place the content onset at the same layer argues L17 is a property of the model, not of a specific task like translation.

The two effects also stack rather than compete. As shown in Figure~\ref{fig:decomp} (left), after the language flip is already complete at L16, \textit{matched} separates from \textit{shuffled} at exactly L17 ($36.1$ vs.\ $31.8$, $p<10^{-3}$).

\paragraph{A late-layer surface effect: \textit{mean} beats \textit{matched}.} \label{sec:mean_vs_matched}
Figure~\ref{fig:decomp} (left) shows that from L20 onward, \textit{mean} overtakes \textit{matched} on $\Delta_{\textsc{high}}$ ($36.3$ vs.\ $31.5$ at L32, paired gap $-4.81$, $p<10^{-3}$), suggesting that the matched donor contains instance-specific content that may compete with the model's own encoding. However, xCOMET-XL does not reproduce that ordering, putting \textit{matched} slightly ahead at L20 ($0.21$ vs.\ $0.19$) and level at L32. We therefore regard this effect as one of surface form: the matched donor may change wording away from the reference without reducing adequacy. We further explore and explain this effect using SAEs in \S\ref{sec:feature}.

\begin{figure}[h]
  \centering
  \includegraphics[width=0.9\columnwidth]{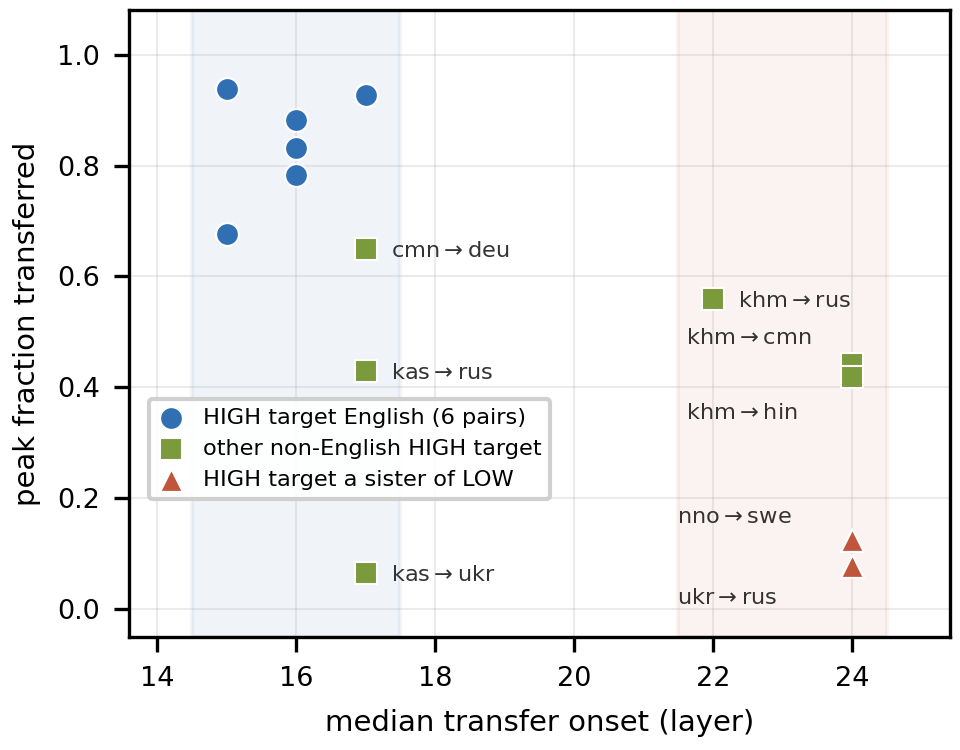}
  \caption{Target-flip onset atlas over all $15$ different-target pairs under \textit{matched} patching. The horizontal axis is the median \textit{onset layer}, computed from the target-exclusive-character discriminator, and the vertical axis is the peak \textit{fraction} of instances whose output fastText identifies as the HIGH target language. Pairs whose HIGH target is high-resource Latin form one cluster (onset L15--17, $.68$--$.94$ flipped); sister targets sit at L24 and near zero; other non-English targets fall between. One pair, \texttt{eng}$\rightarrow$\texttt{kas} vs.\ \texttt{eng}$\rightarrow$\texttt{hin}, is omitted because its discriminator is saturated at baseline: $91.8\%$ of clean \texttt{eng}$\rightarrow$\texttt{kas} outputs are already identified as Hindi.}
  \label{fig.onset_atlas}
\end{figure}

\paragraph{Cross-script and sister targets resolve late.}
\label{non_eng_onset}
Figure~\ref{fig.onset_atlas} plots the flip \textit{onset layer} against the peak \textit{fraction} for all $15$ different-target pairs. When the HIGH direction targets something other than English, both the onset and the magnitude change. Flip onsets mostly\footnote{Pairs with Kashmiri (kas) as target are an exception. With a mean xCOMET-XL of $0.13$, almost anything foreign-shaped overrides kas's weak default, and all four donors switch together at L17 without separation of language and content.} move to L22--24, and the \textit{flip fraction} tracks how distant the two targets are: $0.44$ for khm (Khmer) vs. cmn (Chinese) and $0.42$ for khm vs. hin (Hindi), both cross-script, against only $0.12$ for the nno/swe (Nynorsk/Swedish) sisters and $0.08$ for ukr/rus (Ukrainian/Russian). The English (or high-resource Latin) and non-English targets form two separate clusters regarding the flip \textit{onset layer} and \textit{flip fraction}. An intervention-free logit-lens probe in Appendix~\ref{sec:lens} validates this onset map, corroborating results in~\citet{zeng-etal-2025-converging}.

\section{Feature-level Causality}
\label{sec:feature}

$\S$\ref{decomp_results} reveals a three-stage layer map for LLM multilingual generation, but leaves open two questions: 1) is the language flip carried by a small, nameable set of features, and 2) is the late-layer \textit{mean}-over-\textit{matched} gap really instance-specific content? 

To answer the questions, we encode the LOW and HIGH last-token residuals with Gemma Scope 2 residual-stream SAEs (width $262$k). The reconstruction is faithful, as the mean $1 - \lVert\hat{\mathbf{h}}-\mathbf{h}\rVert/\lVert\mathbf{h}\rVert$ is from $0.96$ at L13 to $0.90$ at L28, with $L_0 \approx 100$--$150$ active features. Features are ranked per pair and standardised across instances. To evaluate how much of the flip direction $\mathbf{r} = \bar{\mathbf{h}}_{\textsc{high}} - \bar{\mathbf{h}}_{\textsc{low}}$ is spanned by the top-$k$ SAE features, we compute the cumulative $R^2$ as a function of $k$. Let $X_k \in \mathbb{R}^{d \times k}$ be the matrix whose columns are the decoder directions $\mathbf{d}_f$ of the top-$k$ features. We define $R^2_k$:
\begin{equation} \label{eq.rsquared}
R^2_k = 1-\frac{\min_{\boldsymbol{\beta} \in \mathbb{R}^{k}} \lVert \mathbf{r}-X_k \boldsymbol{\beta} \rVert^2}{\lVert \mathbf{r} \rVert^2},
\end{equation}
which measures the fraction of variance of $\mathbf{r}$ captured by the linear subspace spanned by these $k$ directions. $R^2_k = 1$ suggests the subspace contains $\mathbf{r}$ exactly, whereas $R^2_k = 0$ indicates orthogonality. We fit $\boldsymbol{\beta}$ by least squares rather than fixing it to the features' raw activation differences, so scale mismatches cannot understate what the set spans.

\paragraph{The English flip is not sparse.}
%Figure~\ref{fig.sae_flip} (left) in Appendix~\ref{sec:add_results} shows $R^2_{50}$ per layer for the six English-HIGH pairs. 
$R^2_{50}$ at L13 sits at $0.80 \pm 0.14$, but drops to $0.48 \pm 0.27$ at L15 and $0.54 \pm 0.11$ at L16, and it recovers to $0.76 \pm 0.11$ at L17 (Figure~\ref{fig.sae_flip} left, Appendix~\ref{sec:add_results}). The dip coincides exactly with the layers where the English flip is causal. It is not a matter of the amount of features, as doubling the budget to $k{=}100$ raises L15 only from $0.48$ to $0.54$. This indicates that the English-flip direction is represented \emph{least} sparsely by the top$-k$ SAE features at this layer.

Since the decoder directions $\mathbf{d}_f$ live in the model's own residual space, any set of features can be turned back into a vector and added to or subtracted from an activation for feature steering~\citep{al-ghussin-etal-2026-multilingual,zhao-etal-2026-denoising}. Let $S_{\textsc{low}}$ and $S_{\textsc{high}}$ denote the selected feature sets associated with the LOW and HIGH pairs, and let $w_f$ denote the steering weight assigned to feature $f$. For an activation $\mathbf{h}$, we define three interventions:
\begin{align}
\Delta\mathbf{h}_{\text{ablate}}
&=
-\sum_{f\in S_{\textsc{low}}}
a_f(\mathbf{h}) \mathbf{d}_f,
\\
\Delta\mathbf{h}_{\text{add}}
&=
\sum_{f\in S_{\textsc{high}}}
w_f \mathbf{d}_f,
\\
\Delta\mathbf{h}_{\text{both}}
&=
\Delta\mathbf{h}_{\text{ablate}}
+
\Delta\mathbf{h}_{\text{add}}.
\end{align}
The intervened residual state is then
\begin{equation}
\mathbf{h}'=\mathbf{h}+\Delta\mathbf{h}.
\end{equation}
Thus, \textit{ablate} removes the currently active LOW-associated features, \textit{add} injects the selected HIGH-associated features, and \textit{both} performs both operations. We evaluate these interventions as additional conditions in the same layer-wise sweep used for activation patching. Each feature set is manually verified in relation to the target languages on the max-activating-feature dashboard (Figure~\ref{fig.dashboard}) before used for patching. This leads to $3$--$7$ features per layer. Our results show that on every pair whose HIGH target is English, \textit{matched}, \textit{mean} and \textit{mean\_diff} put $76$--$97\%$ of outputs into English at the onset layer, while all three sparse interventions leave the output language at its clean baseline of $0$--$5\%$ (Figure~\ref{fig.patching_flip_English} for \texttt{kas}$\rightarrow$\texttt{cmn} vs.\ \texttt{hin}$\rightarrow$\texttt{eng}). The negative result holds across both SAE widths ($16$k and $262$k), for $k \le 100$, and at steering strength $\alpha = 1$ as well as $\alpha = 4$.

\paragraph{Sister features are sparse, but not last-token causal.}
$R^2_{50}$ at L24--25 is $0.91$--$0.95$, suggesting that the direction that separates sister language pairs including \texttt{nno} from \texttt{swe}, or \texttt{ukr} from \texttt{rus}, is sparsely represented (Figure~\ref{fig.sae_flip} middle and right). It also rules out the obvious objection that our features are badly chosen. However, no identified feature flips the language. As \textit{matched} itself flips only $6$--$7\%$ of instances against a clean baseline of $0$--$3\%$, the sparse interventions are indistinguishable from that (Figure~\ref{fig.patching_flip_sister}). The disjoint pair built from the same sister languages but with English as HIGH target flips $76\%$ of instances at the same layer under the same conditions (Figure~\ref{fig.patching_flip_sister_compare}). It is a sharp contrast that at L24, an English target flips at $76$--$97\%$ and a sister target at noise, when we hold other variables fixed and vary only the granularity of the language distinction. This suggests that sparse representability and last-token causality are independent properties, and the sister language decision has the first, but not the second.

\paragraph{Instance-specific features become relevant at L17.}
To investigate the \textit{mean}-over-\textit{matched} gap ($\S$\ref{sec:mean_vs_matched}), we split each HIGH vector's active features into \emph{shared} (active in $\ge 90\%$ of that pair's instances) and \emph{instance-specific}, and decode each group separately. We find that shared feature mass falls quickly from $0.76$ at L16 to $0.57$ at L17, whereas instance-specific mass gains importance at the same layer as the content-onset layer L17 (Figure~\ref{fig:sae}).

\begin{figure}[t]
  \centering
  \includegraphics[width=0.99\columnwidth]{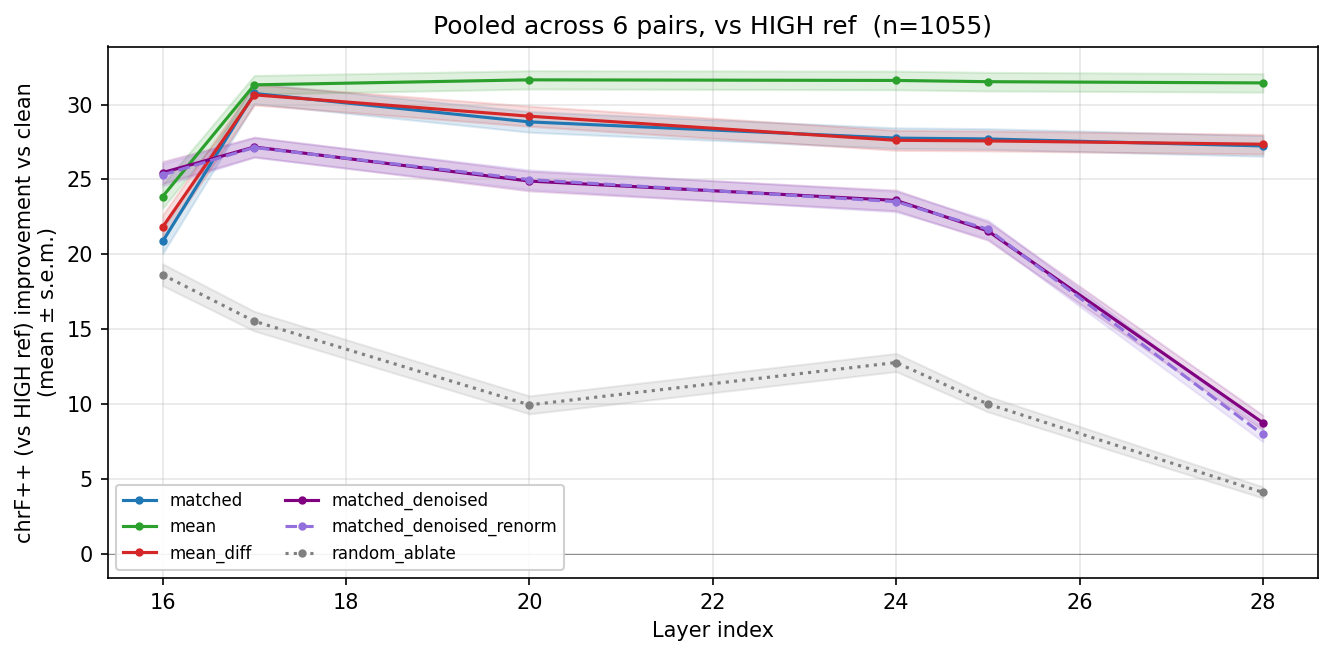}
  \caption{$\Delta_{\textsc{high}}$ when patching the denoised vectors and their two controls ($n{=}1055$), pooled over the six different-target pairs with English as HIGH target.}
  \label{fig:sae_denoised}
\end{figure}

To remove the instance-specific noise, we patch the HIGH vector with its instance-specific feature reconstruction subtracted (\textit{matched\_denoised}), against two other controls: 1) \textit{matched\_denoised\_renorm} which rescales the same direction back to $\lVert\mathbf{h}\rVert$ since subtraction also shrinks the vector, and 2) \textit{random\_ablate} which removes an equal amount of mass along a random equal-count set of active features of that instance. Figure~\ref{fig:sae_denoised} shows the result.

At L16, the denoised vector \emph{beats} the full HIGH vector \textit{matched} by $4.60$ chrF++ ($p<10^{-3}$, $n{=}1055$) and \textit{mean} by $1.60$ ($p<10^{-3}$). It also preserves the language flip, as the fraction of outputs in the HIGH target language is $0.60/0.84/0.79/0.91/0.89/0.86$ for \textit{denoised} against $0.60/0.39/0.78/0.91/0.88/0.57$ for \textit{matched} across six pairs, matching or exceeding it everywhere. Removing an equal mass from random features instead drops the same fractions to $0.49/0.56/0.54/0.68/0.79/0.10$. However, from L17 the ordering reverses. \textit{Denoised} falls below \textit{matched} by $3.58$ at L17, $3.95$ at L20 and $18.48$ at L28 (all $p<10^{-3}$). The two controls show that this is neither an artifact of scale nor of removing mass as such. Renormalising changes nothing at any layer ($+0.15$, $p = 0.26$ at L16; $+0.03$, $p = 0.85$ at L17; $-0.10$, $p = 0.67$ at L20), which suggests the collapse is not the norm damage. Removing the same mass from random features is far worse everywhere ($+6.8$ to $+14.9$ in favour of \textit{denoised}). Therefore, it is the identity of the removed features that carries the effect, not their quantity. This indicates that the shared/instance-specific split is functionally real, and the layer at which instance-specific features become load-bearing is L17---the content onset layer in \S\ref{sec:content_transfer}.

If the late-layer \textit{mean}-over-\textit{matched} gap were caused by interference of instance-specific content with the model’s own encoding, removing that content should have moved \textit{matched} toward \textit{mean}. However, \textit{denoised} moves sharply away from both, suggesting no content interference there.

\section{Cross-model Generalisation} \label{generalization}

\begin{table}[t]
\centering
\resizebox{0.99\linewidth}{!}{%
\begin{tabular}{@{}lccc@{}}
\toprule
 & \textbf{Gemma-3} & \textbf{Qwen3.5} & \textbf{Llama-3.1} \\
 & \textbf{4B} & \textbf{9B} & \textbf{8B} \\
\midrule
Layers / attention & 34 / full & 32 / hybrid & 32 / full \\
English flip onset & 16 ($.47$) & 18 ($.56$) & 13 ($.41$) \\
Content onset & 17 ($.50$) & 19 ($.59$) & 14 ($.44$) \\
Target onset & 24 ($.71$) & 28 ($.88$) & 24 ($.75$) \\
\midrule
\textit{mat.}$-$\textit{shuf.} @ flip & $-0.02^{\dagger}$ & $+0.95^{\dagger}$ & $+0.74^{\ddagger}$ \\
\textit{mat.}$-$\textit{shuf.} @ content & $+4.43$ & $+3.63$ & $+2.06$ \\
\textit{mean}$ \ge $\textit{mat.} @ late & $+4.81$ & $+8.36$ & $+1.94$ \\
\midrule
Belebele clean / mat. & $.10$ / $.87$ & $.02$ / $.99$ & $.02$ / $.98$ \\
Belebele \textit{mat.} / \textit{shuf.} & $.87$ / $.27$ & $.99$ / $.25$ & $.98$ / $.23$ \\
\midrule
SAE recon.\ fidelity & $.90$--$.96$ & $.45$--$.73$ & $.26$--$.55$ \\
Sparse flip works? & no & no & n/a \\
\bottomrule
\end{tabular}%
}
\caption{Replication across three models. Onsets are layer indices with relative depth in parentheses. \emph{Target onset} is the late-layer resolution of the specific target language from the two sister pairs (\texttt{eng}$\rightarrow$\texttt{nno} vs.\ \texttt{eng}$\rightarrow$\texttt{swe}, \texttt{eng}$\rightarrow$\texttt{ukr} vs.\ \texttt{eng}$\rightarrow$\texttt{rus}). chrF++ differences are paired means over pooled instances, all $p<10^{-3}$ except $^{\dagger}$, which are null ($p = 0.96$ and $p = 0.12$). $^{\ddagger}$ is significant ($p<10^{-3}$) but is $1.6\%$ of the $46.6$ chrF++ effect at that layer, against $13\%$ one layer later. Belebele rows are accuracies at the final layer. SAE reconstruction fidelity is $1-\lVert\hat{\mathbf{h}}-\mathbf{h}\rVert/\lVert\mathbf{h}\rVert$ over the swept layers. ``n/a'' marks a sparse-flip test which we could not run because fidelity is too low.}
\label{tab:crossmodel}
\end{table}

We re-ran the whole pipeline on \texttt{Qwen3.5-9B-Base} and \texttt{Llama-3.1-8B}, which differ from Gemma in scale, tokenizer family (BPE vs.\ SentencePiece)~\citep{kudo-richardson-2018-sentencepiece}, SAE family (TopK) and attention design (hybrid for Qwen)~\citep{Qwennext25}. The detailed experimental setups are described in Appendix~\ref{sec:qwen_llama}.

\paragraph{The patching-level account transfers.} Table~\ref{tab:crossmodel} summarises the results. We reproduce the three-stage pattern in \S\ref{decomp_results} for both Qwen and Llama regardless of their differences. In all three models, the content channel opens immediately after the English flip, although the depth at which this happens does not hold constant across models. Belebele shows the sharpest contrast as \textit{matched} rises quickly after the onset layer while \textit{shuffled} plateaus at about $0.25$ for all three models.

For Qwen, the English flip starts two layers later than Gemma, and the paired \textit{matched}$-$\textit{shuffled} difference at the flip layer is not significant ($+0.95$, $p = 0.12$). However, all four donors switch together at L18, followed by the content onset at L19. The late-layer \textit{mean}-over-\textit{matched} effect reappears and is larger than in Gemma ($\textit{mean}-\textit{matched} = 8.36$ chrF++ at L28, $p<10^{-3}$). 

The English flip for Llama, however, switches on at L13, three layers earlier than Gemma.  The donor's identity for Llama makes a small difference ($\textit{matched}-\textit{shuffled}=+0.74$ chrF++, $p<10^{-3}$) at the flip layer, whereas in Gemma and Qwen there is no measurable difference. The language transitions here are ramps rather than steps, as we see $\Delta_{\textsc{high}}$ climbs over L10-13, instead of jumping at a single layer. So the two stages are less cleanly separated. We read Llama as confirming the transfer ordering while noting that the layer-sharpness of the transition is model-specific. %What separates them is the size of the effect rather than its presence: donor identity accounts for $1.6\%$ of the flip at L13, $3.3\%$ at L14 and $13\%$ at L15, the same fraction Gemma reaches at its own content onset. 

\paragraph{The feature-level account transfers only in part.} The SAE analyses are bounded by reconstruction fidelity at the last-token position: $0.45$--$0.73$ for Qwen-Scope and $0.26$--$0.55$ for Llama-Scope, against $0.90$--$0.96$ for Gemma Scope 2 (Table~\ref{tab:crossmodel}). On Qwen, the sparse interventions again fail to flip the output into English while the dense donors succeed, but the \textit{denoised} vector is dominated by reconstruction errors. On Llama, no feature is exclusively language-specific. Detailed analyses of SAE features for both models are presented in Appendix~\ref{sec:qwen_sae}. We interpret this as a limit of the instrument rather than a contradiction of the Gemma results, whereas the patching-level account, which needs no SAE, transfers across all three models.

\section{Conclusion}

This paper introduced a donor-controlled activation-patching method to disentangle the content information and language-related signals in multilingual generation. Across three LLMs, we find a consistent ordering in which an intermediate-layer English default emerges first, content information becomes causally available shortly afterward, and the final target language is resolved only at later layers. This pattern suggests that some low-resource generation failures may reflect not only missing knowledge, but also how and when language selection is established within the forward pass. Future work could examine how this internal organization changes when models are further pretrained on low-resource languages, and whether improved multilingual capability corresponds to shifts in the timing, strength, or geometry of these signals.

\section*{Limitations}

Our interventions are deliberately narrow: we patch a single residual-stream vector at the last token position and vary one layer at a time. The resulting layerwise curves therefore identify what information is causally available at that state, but not necessarily where the underlying computation is performed. In particular, negative results should not be interpreted as evidence that a signal is absent from the model. It may instead be carried by other token positions, distributed across components, or require interventions at multiple layers. Our conclusions are thus about the causal role of the final-position residual stream rather than a complete localisation of the multilingual computation.

The feature-level analysis is likewise relative to the sparse dictionaries we study. Failure to reproduce a causal effect with SAE features does not imply that no sparse representation exists, since the result depends on the learned dictionary, its reconstruction fidelity, and the feature-selection procedure. This is especially important for models or layers with poor reconstruction quality, where null results are difficult to interpret. More broadly, the contrastive instances are selected using xCOMET-XL, whose reliability can vary across languages and may inherit biases toward higher-resource settings, although we report complementary metrics when evaluating the resulting interventions.

Finally, our experiments cover three pretrained models and a limited set of language pairs and tasks. We therefore do not claim that the exact layer locations or effect magnitudes generalise to all architectures, scales, or languages. Our central claim concerns the recurring ordering of the observed stages. Extending the analysis to larger models, instruction-tuned checkpoints, and a broader range of low-resource languages would provide a stronger test of this generality.

\section*{Ethical Considerations}

This research relies exclusively on publicly accessible datasets, with all data utilisation adhering to the license agreements specified by~\citet{maillard-etal-2024-findings,bandarkar-etal-2024-belebele,federmann-etal-2022-ntrex}. It is presumed that these repositories contain no sensitive or personally identifiable information. Consequently, their application in this study is deemed to have no significant ethical risks. 

All original ideas, analyses, and content in this paper were created by the authors. AI tools were used only as supportive aids for improving writing quality and assisting with coding tasks. The authors retain full responsibility for the intellectual content, analyses, and conclusions presented in this work.

\section*{Acknowledgments}

This work has received funding from the European Union's Horizon Europe research and innovation programme under the Marie Skłodowska-Curie grant agreement No. 101126636. 

The computations were performed on resources provided through Sigma2---the national research infrastructure provider for high-performance computing and large-scale data storage in Norway. We acknowledge Norway and Sigma2 for awarding this project access to the Olivia supercomputer, through Project nn9851k.

\bibliography{custom,anthology-1,anthology-2}

@inproceedings{ZhangBiao2023,
author = {Zhang, Biao and Haddow, Barry and Birch, Alexandra},
title = {{Prompting large language model for machine translation: a case study}},
year = {2023},
publisher = {JMLR.org},
booktitle = {Proceedings of the 40th International Conference on Machine Learning},
articleno = {1722},
numpages = {19},
location = {Honolulu, Hawaii, USA},
series = {ICML'23},
url = {https://dl.acm.org/doi/abs/10.5555/3618408.3620130}
}

@ARTICLE{Grattafiori2024-ot,
  title         = "The Llama 3 herd of models",
  author        = "Grattafiori, Aaron and Dubey, Abhimanyu and Jauhri, Abhinav
                   and Pandey, Abhinav and Kadian, Abhishek and Al-Dahle, Ahmad
                   and Letman, Aiesha and Mathur, Akhil and Schelten, Alan and
                   Vaughan, Alex and Yang, Amy and Fan, Angela and Goyal,
                   Anirudh and Hartshorn, Anthony and Yang, Aobo and Mitra,
                   Archi and Sravankumar, Archie and Korenev, Artem and
                   Hinsvark, Arthur and Rao, Arun and Zhang, Aston and
                   Rodriguez, Aurelien and Gregerson, Austen and Spataru, Ava
                   and Roziere, Baptiste and Biron, Bethany and Tang, Binh and
                   Chern, Bobbie and Caucheteux, Charlotte and Nayak, Chaya and
                   Bi, Chloe and Marra, Chris and McConnell, Chris and Keller,
                   Christian and Touret, Christophe and Wu, Chunyang and Wong,
                   Corinne and Ferrer, Cristian Canton and Nikolaidis, Cyrus
                   and Allonsius, Damien and Song, Daniel and Pintz, Danielle
                   and Livshits, Danny and Wyatt, Danny and Esiobu, David and
                   Choudhary, Dhruv and Mahajan, Dhruv and Garcia-Olano, Diego
                   and Perino, Diego and Hupkes, Dieuwke and Lakomkin, Egor and
                   AlBadawy, Ehab and Lobanova, Elina and Dinan, Emily and
                   Smith, Eric Michael and Radenovic, Filip and Guzm{\'a}n,
                   Francisco and Zhang, Frank and Synnaeve, Gabriel and Lee,
                   Gabrielle and Anderson, Georgia Lewis and Thattai, Govind
                   and Nail, Graeme and Mialon, Gregoire and Pang, Guan and
                   Cucurell, Guillem and Nguyen, Hailey and Korevaar, Hannah
                   and Xu, Hu and Touvron, Hugo and Zarov, Iliyan and Ibarra,
                   Imanol Arrieta and Kloumann, Isabel and Misra, Ishan and
                   Evtimov, Ivan and Zhang, Jack and Copet, Jade and Lee,
                   Jaewon and Geffert, Jan and Vranes, Jana and Park, Jason and
                   Mahadeokar, Jay and Shah, Jeet and van der Linde, Jelmer and
                   Billock, Jennifer and Hong, Jenny and Lee, Jenya and Fu,
                   Jeremy and Chi, Jianfeng and Huang, Jianyu and Liu, Jiawen
                   and Wang, Jie and Yu, Jiecao and Bitton, Joanna and Spisak,
                   Joe and Park, Jongsoo and Rocca, Joseph and Johnstun, Joshua
                   and Saxe, Joshua and Jia, Junteng and Alwala, Kalyan Vasuden
                   and Prasad, Karthik and Upasani, Kartikeya and Plawiak, Kate
                   and Li, Ke and Heafield, Kenneth and Stone, Kevin and
                   El-Arini, Khalid and Iyer, Krithika and Malik, Kshitiz and
                   Chiu, Kuenley and Bhalla, Kunal and Lakhotia, Kushal and
                   Rantala-Yeary, Lauren and van der Maaten, Laurens and Chen,
                   Lawrence and Tan, Liang and Jenkins, Liz and Martin, Louis
                   and Madaan, Lovish and Malo, Lubo and Blecher, Lukas and
                   Landzaat, Lukas and de Oliveira, Luke and Muzzi, Madeline
                   and Pasupuleti, Mahesh and Singh, Mannat and Paluri, Manohar
                   and Kardas, Marcin and Tsimpoukelli, Maria and Oldham,
                   Mathew and Rita, Mathieu and Pavlova, Maya and Kambadur,
                   Melanie and Lewis, Mike and Si, Min and Singh, Mitesh Kumar
                   and Hassan, Mona and Goyal, Naman and Torabi, Narjes and
                   Bashlykov, Nikolay and Bogoychev, Nikolay and Chatterji,
                   Niladri and Zhang, Ning and Duchenne, Olivier and {\c
                   C}elebi, Onur and Alrassy, Patrick and Zhang, Pengchuan and
                   Li, Pengwei and Vasic, Petar and Weng, Peter and Bhargava,
                   Prajjwal and Dubal, Pratik and Krishnan, Praveen and Koura,
                   Punit Singh and Xu, Puxin and He, Qing and Dong, Qingxiao
                   and Srinivasan, Ragavan and Ganapathy, Raj and Calderer,
                   Ramon and Cabral, Ricardo Silveira and Stojnic, Robert and
                   Raileanu, Roberta and Maheswari, Rohan and Girdhar, Rohit
                   and Patel, Rohit and Sauvestre, Romain and Polidoro, Ronnie
                   and Sumbaly, Roshan and Taylor, Ross and Silva, Ruan and
                   Hou, Rui and Wang, Rui and Hosseini, Saghar and
                   Chennabasappa, Sahana and Singh, Sanjay and Bell, Sean and
                   Kim, Seohyun Sonia and Edunov, Sergey and Nie, Shaoliang and
                   Narang, Sharan and Raparthy, Sharath and Shen, Sheng and
                   Wan, Shengye and Bhosale, Shruti and Zhang, Shun and
                   Vandenhende, Simon and Batra, Soumya and Whitman, Spencer
                   and Sootla, Sten and Collot, Stephane and Gururangan, Suchin
                   and Borodinsky, Sydney and Herman, Tamar and Fowler, Tara
                   and Sheasha, Tarek and Georgiou, Thomas and Scialom, Thomas
                   and Speckbacher, Tobias and Mihaylov, Todor and Xiao, Tong
                   and Karn, Ujjwal and Goswami, Vedanuj and Gupta, Vibhor and
                   Ramanathan, Vignesh and Kerkez, Viktor and Gonguet, Vincent
                   and Do, Virginie and Vogeti, Vish and Albiero, V{\'\i}tor
                   and Petrovic, Vladan and Chu, Weiwei and Xiong, Wenhan and
                   Fu, Wenyin and Meers, Whitney and Martinet, Xavier and Wang,
                   Xiaodong and Wang, Xiaofang and Tan, Xiaoqing Ellen and Xia,
                   Xide and Xie, Xinfeng and Jia, Xuchao and Wang, Xuewei and
                   Goldschlag, Yaelle and Gaur, Yashesh and Babaei, Yasmine and
                   Wen, Yi and Song, Yiwen and Zhang, Yuchen and Li, Yue and
                   Mao, Yuning and Coudert, Zacharie Delpierre and Yan, Zheng
                   and Chen, Zhengxing and Papakipos, Zoe and Singh, Aaditya
                   and Srivastava, Aayushi and Jain, Abha and Kelsey, Adam and
                   Shajnfeld, Adam and Gangidi, Adithya and Victoria, Adolfo
                   and Goldstand, Ahuva and Menon, Ajay and Sharma, Ajay and
                   Boesenberg, Alex and Baevski, Alexei and Feinstein, Allie
                   and Kallet, Amanda and Sangani, Amit and Teo, Amos and
                   Yunus, Anam and Lupu, Andrei and Alvarado, Andres and
                   Caples, Andrew and Gu, Andrew and Ho, Andrew and Poulton,
                   Andrew and Ryan, Andrew and Ramchandani, Ankit and Dong,
                   Annie and Franco, Annie and Goyal, Anuj and Saraf, Aparajita
                   and Chowdhury, Arkabandhu and Gabriel, Ashley and Bharambe,
                   Ashwin and Eisenman, Assaf and Yazdan, Azadeh and James,
                   Beau and Maurer, Ben and Leonhardi, Benjamin and Huang,
                   Bernie and Loyd, Beth and De Paola, Beto and Paranjape,
                   Bhargavi and Liu, Bing and Wu, Bo and Ni, Boyu and Hancock,
                   Braden and Wasti, Bram and Spence, Brandon and Stojkovic,
                   Brani and Gamido, Brian and Montalvo, Britt and Parker, Carl
                   and Burton, Carly and Mejia, Catalina and Liu, Ce and Wang,
                   Changhan and Kim, Changkyu and Zhou, Chao and Hu, Chester
                   and Chu, Ching-Hsiang and Cai, Chris and Tindal, Chris and
                   Feichtenhofer, Christoph and Gao, Cynthia and Civin, Damon
                   and Beaty, Dana and Kreymer, Daniel and Li, Daniel and
                   Adkins, David and Xu, David and Testuggine, Davide and
                   David, Delia and Parikh, Devi and Liskovich, Diana and Foss,
                   Didem and Wang, Dingkang and Le, Duc and Holland, Dustin and
                   Dowling, Edward and Jamil, Eissa and Montgomery, Elaine and
                   Presani, Eleonora and Hahn, Emily and Wood, Emily and Le,
                   Eric-Tuan and Brinkman, Erik and Arcaute, Esteban and
                   Dunbar, Evan and Smothers, Evan and Sun, Fei and Kreuk,
                   Felix and Tian, Feng and Kokkinos, Filippos and Ozgenel,
                   Firat and Caggioni, Francesco and Kanayet, Frank and Seide,
                   Frank and Florez, Gabriela Medina and Schwarz, Gabriella and
                   Badeer, Gada and Swee, Georgia and Halpern, Gil and Herman,
                   Grant and Sizov, Grigory and Zhang, Guangyi and
                   Lakshminarayanan, Guna and Inan, Hakan and Shojanazeri,
                   Hamid and Zou, Han and Wang, Hannah and Zha, Hanwen and
                   Habeeb, Haroun and Rudolph, Harrison and Suk, Helen and
                   Aspegren, Henry and Goldman, Hunter and Zhan, Hongyuan and
                   Damlaj, Ibrahim and Molybog, Igor and Tufanov, Igor and
                   Leontiadis, Ilias and Veliche, Irina-Elena and Gat, Itai and
                   Weissman, Jake and Geboski, James and Kohli, James and Lam,
                   Janice and Asher, Japhet and Gaya, Jean-Baptiste and Marcus,
                   Jeff and Tang, Jeff and Chan, Jennifer and Zhen, Jenny and
                   Reizenstein, Jeremy and Teboul, Jeremy and Zhong, Jessica
                   and Jin, Jian and Yang, Jingyi and Cummings, Joe and
                   Carvill, Jon and Shepard, Jon and McPhie, Jonathan and
                   Torres, Jonathan and Ginsburg, Josh and Wang, Junjie and Wu,
                   Kai and Kam, Hou U and Saxena, Karan and Khandelwal,
                   Kartikay and Zand, Katayoun and Matosich, Kathy and
                   Veeraraghavan, Kaushik and Michelena, Kelly and Li, Keqian
                   and Jagadeesh, Kiran and Huang, Kun and Chawla, Kunal and
                   Huang, Kyle and Chen, Lailin and Garg, Lakshya and Lavender,
                   A and Silva, Leandro and Bell, Lee and Zhang, Lei and Guo,
                   Liangpeng and Yu, Licheng and Moshkovich, Liron and
                   Wehrstedt, Luca and Khabsa, Madian and Avalani, Manav and
                   Bhatt, Manish and Mankus, Martynas and Hasson, Matan and
                   Lennie, Matthew and Reso, Matthias and Groshev, Maxim and
                   Naumov, Maxim and Lathi, Maya and Keneally, Meghan and Liu,
                   Miao and Seltzer, Michael L and Valko, Michal and Restrepo,
                   Michelle and Patel, Mihir and Vyatskov, Mik and Samvelyan,
                   Mikayel and Clark, Mike and Macey, Mike and Wang, Mike and
                   Hermoso, Miquel Jubert and Metanat, Mo and Rastegari,
                   Mohammad and Bansal, Munish and Santhanam, Nandhini and
                   Parks, Natascha and White, Natasha and Bawa, Navyata and
                   Singhal, Nayan and Egebo, Nick and Usunier, Nicolas and
                   Mehta, Nikhil and Laptev, Nikolay Pavlovich and Dong, Ning
                   and Cheng, Norman and Chernoguz, Oleg and Hart, Olivia and
                   Salpekar, Omkar and Kalinli, Ozlem and Kent, Parkin and
                   Parekh, Parth and Saab, Paul and Balaji, Pavan and Rittner,
                   Pedro and Bontrager, Philip and Roux, Pierre and Dollar,
                   Piotr and Zvyagina, Polina and Ratanchandani, Prashant and
                   Yuvraj, Pritish and Liang, Qian and Alao, Rachad and
                   Rodriguez, Rachel and Ayub, Rafi and Murthy, Raghotham and
                   Nayani, Raghu and Mitra, Rahul and Parthasarathy,
                   Rangaprabhu and Li, Raymond and Hogan, Rebekkah and Battey,
                   Robin and Wang, Rocky and Howes, Russ and Rinott, Ruty and
                   Mehta, Sachin and Siby, Sachin and Bondu, Sai Jayesh and
                   Datta, Samyak and Chugh, Sara and Hunt, Sara and Dhillon,
                   Sargun and Sidorov, Sasha and Pan, Satadru and Mahajan,
                   Saurabh and Verma, Saurabh and Yamamoto, Seiji and
                   Ramaswamy, Sharadh and Lindsay, Shaun and Lindsay, Shaun and
                   Feng, Sheng and Lin, Shenghao and Zha, Shengxin Cindy and
                   Patil, Shishir and Shankar, Shiva and Zhang, Shuqiang and
                   Zhang, Shuqiang and Wang, Sinong and Agarwal, Sneha and
                   Sajuyigbe, Soji and Chintala, Soumith and Max, Stephanie and
                   Chen, Stephen and Kehoe, Steve and Satterfield, Steve and
                   Govindaprasad, Sudarshan and Gupta, Sumit and Deng, Summer
                   and Cho, Sungmin and Virk, Sunny and Subramanian, Suraj and
                   Choudhury, Sy and Goldman, Sydney and Remez, Tal and Glaser,
                   Tamar and Best, Tamara and Koehler, Thilo and Robinson,
                   Thomas and Li, Tianhe and Zhang, Tianjun and Matthews, Tim
                   and Chou, Timothy and Shaked, Tzook and Vontimitta, Varun
                   and Ajayi, Victoria and Montanez, Victoria and Mohan, Vijai
                   and Kumar, Vinay Satish and Mangla, Vishal and Ionescu, Vlad
                   and Poenaru, Vlad and Mihailescu, Vlad Tiberiu and Ivanov,
                   Vladimir and Li, Wei and Wang, Wenchen and Jiang, Wenwen and
                   Bouaziz, Wes and Constable, Will and Tang, Xiaocheng and Wu,
                   Xiaojian and Wang, Xiaolan and Wu, Xilun and Gao, Xinbo and
                   Kleinman, Yaniv and Chen, Yanjun and Hu, Ye and Jia, Ye and
                   Qi, Ye and Li, Yenda and Zhang, Yilin and Zhang, Ying and
                   Adi, Yossi and Nam, Youngjin and Wang, Yu and Zhao, Yu and
                   Hao, Yuchen and Qian, Yundi and Li, Yunlu and He, Yuzi and
                   Rait, Zach and DeVito, Zachary and Rosnbrick, Zef and Wen,
                   Zhaoduo and Yang, Zhenyu and Zhao, Zhiwei and Ma, Zhiyu",
  month         =  jul,
  year          =  2024,
  copyright     = "http://arxiv.org/licenses/nonexclusive-distrib/1.0/",
  archivePrefix = "arXiv",
  primaryClass  = "cs.AI",
  journal       = "arXiv preprint",
  url           = "https://arxiv.org/abs/2407.21783",
  eprint        = "2407.21783"
}

@ARTICLE{Gemma_Team2025-wv,
  title         = "Gemma 3 Technical Report",
  author        = "{Gemma Team} and Kamath, Aishwarya and Ferret, Johan and Pathak,
                   Shreya and Vieillard, Nino and Merhej, Ramona and Perrin,
                   Sarah and Matejovicova, Tatiana and Ram{\'e}, Alexandre and
                   Rivi{\`e}re, Morgane and Rouillard, Louis and Mesnard,
                   Thomas and Cideron, Geoffrey and Grill, Jean-Bastien and
                   Ramos, Sabela and Yvinec, Edouard and Casbon, Michelle and
                   Pot, Etienne and Penchev, Ivo and Liu, Ga{\"e}l and Visin,
                   Francesco and Kenealy, Kathleen and Beyer, Lucas and Zhai,
                   Xiaohai and Tsitsulin, Anton and Busa-Fekete, Robert and
                   Feng, Alex and Sachdeva, Noveen and Coleman, Benjamin and
                   Gao, Yi and Mustafa, Basil and Barr, Iain and Parisotto,
                   Emilio and Tian, David and Eyal, Matan and Cherry, Colin and
                   Peter, Jan-Thorsten and Sinopalnikov, Danila and
                   Bhupatiraju, Surya and Agarwal, Rishabh and Kazemi, Mehran
                   and Malkin, Dan and Kumar, Ravin and Vilar, David and
                   Brusilovsky, Idan and Luo, Jiaming and Steiner, Andreas and
                   Friesen, Abe and Sharma, Abhanshu and Sharma, Abheesht and
                   Gilady, Adi Mayrav and Goedeckemeyer, Adrian and Saade, Alaa
                   and Feng, Alex and Kolesnikov, Alexander and Bendebury,
                   Alexei and Abdagic, Alvin and Vadi, Amit and Gy{\"o}rgy,
                   Andr{\'a}s and Pinto, Andr{\'e} Susano and Das, Anil and
                   Bapna, Ankur and Miech, Antoine and Yang, Antoine and
                   Paterson, Antonia and Shenoy, Ashish and Chakrabarti, Ayan
                   and Piot, Bilal and Wu, Bo and Shahriari, Bobak and Petrini,
                   Bryce and Chen, Charlie and Le Lan, Charline and
                   Choquette-Choo, Christopher A and Carey, C J and Brick,
                   Cormac and Deutsch, Daniel and Eisenbud, Danielle and
                   Cattle, Dee and Cheng, Derek and Paparas, Dimitris and
                   Sreepathihalli, Divyashree Shivakumar and Reid, Doug and
                   Tran, Dustin and Zelle, Dustin and Noland, Eric and
                   Huizenga, Erwin and Kharitonov, Eugene and Liu, Frederick
                   and Amirkhanyan, Gagik and Cameron, Glenn and Hashemi, Hadi
                   and Klimczak-Pluci{\'n}ska, Hanna and Singh, Harman and
                   Mehta, Harsh and Lehri, Harshal Tushar and Hazimeh, Hussein
                   and Ballantyne, Ian and Szpektor, Idan and Nardini, Ivan and
                   Pouget-Abadie, Jean and Chan, Jetha and Stanton, Joe and
                   Wieting, John and Lai, Jonathan and Orbay, Jordi and
                   Fernandez, Joseph and Newlan, Josh and Ji, Ju-Yeong and
                   Singh, Jyotinder and Black, Kat and Yu, Kathy and Hui, Kevin
                   and Vodrahalli, Kiran and Greff, Klaus and Qiu, Linhai and
                   Valentine, Marcella and Coelho, Marina and Ritter, Marvin
                   and Hoffman, Matt and Watson, Matthew and Chaturvedi, Mayank
                   and Moynihan, Michael and Ma, Min and Babar, Nabila and Noy,
                   Natasha and Byrd, Nathan and Roy, Nick and Momchev, Nikola
                   and Chauhan, Nilay and Sachdeva, Noveen and Bunyan, Oskar
                   and Botarda, Pankil and Caron, Paul and Rubenstein, Paul
                   Kishan and Culliton, Phil and Schmid, Philipp and Sessa,
                   Pier Giuseppe and Xu, Pingmei and Stanczyk, Piotr and Tafti,
                   Pouya and Shivanna, Rakesh and Wu, Renjie and Pan, Renke and
                   Rokni, Reza and Willoughby, Rob and Vallu, Rohith and
                   Mullins, Ryan and Jerome, Sammy and Smoot, Sara and Girgin,
                   Sertan and Iqbal, Shariq and Reddy, Shashir and Sheth,
                   Shruti and P{\~o}der, Siim and Bhatnagar, Sijal and Panyam,
                   Sindhu Raghuram and Eiger, Sivan and Zhang, Susan and Liu,
                   Tianqi and Yacovone, Trevor and Liechty, Tyler and Kalra,
                   Uday and Evci, Utku and Misra, Vedant and Roseberry, Vincent
                   and Feinberg, Vlad and Kolesnikov, Vlad and Han, Woohyun and
                   Kwon, Woosuk and Chen, Xi and Chow, Yinlam and Zhu, Yuvein
                   and Wei, Zichuan and Egyed, Zoltan and Cotruta, Victor and
                   Giang, Minh and Kirk, Phoebe and Rao, Anand and Black, Kat
                   and Babar, Nabila and Lo, Jessica and Moreira, Erica and
                   Martins, Luiz Gustavo and Sanseviero, Omar and Gonzalez,
                   Lucas and Gleicher, Zach and Warkentin, Tris and Mirrokni,
                   Vahab and Senter, Evan and Collins, Eli and Barral, Joelle
                   and Ghahramani, Zoubin and Hadsell, Raia and Matias, Yossi
                   and Sculley, D and Petrov, Slav and Fiedel, Noah and
                   Shazeer, Noam and Vinyals, Oriol and Dean, Jeff and
                   Hassabis, Demis and Kavukcuoglu, Koray and Farabet, Clement
                   and Buchatskaya, Elena and Alayrac, Jean-Baptiste and Anil,
                   Rohan and Lepikhin, Dmitry and Borgeaud, Sebastian and
                   Bachem, Olivier and Joulin, Armand and Andreev, Alek and
                   Hardin, Cassidy and Dadashi, Robert and Hussenot,
                   L{\'e}onard",
  month         =  mar,
  year          =  2025,
  copyright     = "http://creativecommons.org/licenses/by/4.0/",
  archivePrefix = "arXiv",
  primaryClass  = "cs.CL",
  journal       = "arXiv preprint",
  url           = "https://arxiv.org/abs/2503.19786",
  eprint        = "2503.19786"
}

@misc{qwen3.5,
    title  = {{Qwen3.5}: Towards Native Multimodal Agents},
    author = {{Qwen Team}},
    month  = {February},
    year   = {2026},
    url    = {https://qwen.ai/blog?id=qwen3.5}
}

@misc{eval-harness,
  author       = {Gao, Leo and Tow, Jonathan and Abbasi, Baber and Biderman, Stella and Black, Sid and DiPofi, Anthony and Foster, Charles and Golding, Laurence and Hsu, Jeffrey and Le Noac'h, Alain and Li, Haonan and McDonell, Kyle and Muennighoff, Niklas and Ociepa, Chris and Phang, Jason and Reynolds, Laria and Schoelkopf, Hailey and Skowron, Aviya and Sutawika, Lintang and Tang, Eric and Thite, Anish and Wang, Ben and Wang, Kevin and Zou, Andy},
  title        = {The Language Model Evaluation Harness},
  month        = 07,
  year         = 2024,
  publisher    = {Zenodo},
  version      = {v0.4.3},
  doi          = {10.5281/zenodo.12608602},
  url          = {https://zenodo.org/records/12608602}
}

@ARTICLE{Fiotto-Kaufman2024-dj,
  title         = "{NNsight} and {NDIF}: Democratizing access to open-weight
                   foundation model internals",
  author        = "Fiotto-Kaufman, Jaden and Loftus, Alexander R and Todd, Eric
                   and Brinkmann, Jannik and Pal, Koyena and Troitskii, Dmitrii
                   and Ripa, Michael and Belfki, Adam and Rager, Can and Juang,
                   Caden and Mueller, Aaron and Marks, Samuel and Sharma, Arnab
                   Sen and Lucchetti, Francesca and Prakash, Nikhil and
                   Brodley, Carla and Guha, Arjun and Bell, Jonathan and
                   Wallace, Byron C and Bau, David",
  year          =  {2024},
  copyright     = "http://creativecommons.org/licenses/by/4.0/",
  archivePrefix = "arXiv",
  primaryClass  = "cs.LG",
  eprint        = "2407.14561",
  url           ={https://arxiv.org/abs/2407.14561}, 
  journal       ="arXiv preprint"
}

@misc{gemmascope2,
    title  = {{Gemma Scope 2 - Technical Paper}},
    author = {McDougall, Callum and Conmy, Arthur and Kramár, János and Lieberum, Tom and Rajamanoharan, Senthooran and Nanda, Neel},
    month  = {September},
    year   = {2025},
    url    = {https://storage.googleapis.com/deepmind-media/DeepMind.com/Blog/gemma-scope-2-helping-the-ai-safety-community-deepen-understanding-of-complex-language-model-behavior/Gemma_Scope_2_Technical_Paper.pdf}
}

@inproceedings{kwon2023vllm,
author = {Kwon, Woosuk and Li, Zhuohan and Zhuang, Siyuan and Sheng, Ying and Zheng, Lianmin and Yu, Cody Hao and Gonzalez, Joseph and Zhang, Hao and Stoica, Ion},
title = {Efficient Memory Management for Large Language Model Serving with {P}aged{A}ttention},
year = {2023},
isbn = {9798400702297},
publisher = {Association for Computing Machinery},
address = {New York, NY, USA},
url = {https://doi.org/10.1145/3600006.3613165},
doi = {10.1145/3600006.3613165},
booktitle = {Proceedings of the 29th Symposium on Operating Systems Principles},
pages = {611–626},
numpages = {16},
location = {Koblenz, Germany},
series = {SOSP '23}
}

@misc{templeton2024scaling,
  title  = {Scaling Monosemanticity: Extracting Interpretable Features from {C}laude 3 {S}onnet},
  author = {Templeton, Adly and Conerly, Tom and Marcus, Jonathan and Lindsey, Jack and Bricken, Trenton and Chen, Brian and Pearce, Adam and Citro, Craig and Ameisen, Emmanuel and Jones, Andy and Cunningham, Hoagy and Turner, Nicholas L. and McDougall, Callum and MacDiarmid, Monte and Tamkin, Alex and Durmus, Esin and Hume, Tristan and Mosconi, Francesco and Freeman, C. Daniel and Sumers, Theodore R. and Rees, Edward and Batson, Joshua and Jermyn, Adam and Carter, Shan and Olah, Chris and Henighan, Tom},
  year   = {2024},
  note   = {Transformer Circuits Thread},
  url    = {https://transformer-circuits.pub/2024/scaling-monosemanticity/}
}

@misc{bricken2023monosemanticity,
  title  = {Towards Monosemanticity: Decomposing Language Models With Dictionary Learning},
  author = {Bricken, Trenton and Templeton, Adly and Batson, Joshua and Chen, Brian and Jermyn, Adam and Conerly, Tom and Turner, Nicholas L. and Anil, Cem and Denison, Carson and Askell, Amanda and Lasenby, Robert and Wu, Yifan and Kravec, Shauna and Schiefer, Nicholas and Maxwell, Tim and Joseph, Nicholas and Tamkin, Alex and Nguyen, Karina and McLean, Brayden and Burke, Josiah E. and Hume, Tristan and Carter, Shan and Henighan, Tom and Olah, Chris},
  year   = {2023},
  note   = {Transformer Circuits Thread},
  url    = {https://transformer-circuits.pub/2023/monosemantic-features/}
}

@inproceedings{
huben2024sparse,
title={Sparse Autoencoders Find Highly Interpretable Features in Language Models},
author={Robert Huben and Hoagy Cunningham and Logan Riggs Smith and Aidan Ewart and Lee Sharkey},
booktitle={The Twelfth International Conference on Learning Representations},
year={2024},
url={https://openreview.net/forum?id=F76bwRSLeK}
}

@inproceedings{
todd2024function,
title={Function Vectors in Large Language Models},
author={Eric Todd and Millicent Li and Arnab Sen Sharma and Aaron Mueller and Byron C Wallace and David Bau},
booktitle={The Twelfth International Conference on Learning Representations},
year={2024},
url={https://openreview.net/forum?id=AwyxtyMwaG}
}

@inproceedings{meng2022,
author = {Meng, Kevin and Bau, David and Andonian, Alex and Belinkov, Yonatan},
title = {Locating and editing factual associations in {GPT}},
year = {2022},
isbn = {9781713871088},
publisher = {Curran Associates Inc.},
address = {Red Hook, NY, USA},
booktitle = {Proceedings of the 36th International Conference on Neural Information Processing Systems},
articleno = {1262},
numpages = {14},
location = {New Orleans, LA, USA},
series = {NIPS '22},
url = {https://dl.acm.org/doi/10.5555/3600270.3601532}
}

@ARTICLE{Deng2026-yx,
  title         = "{Qwen-Scope}: Turning sparse features into development tools
                   for large language models",
  author        = "Deng, Boyi and Wang, Xu and Wang, Yaoning and Wan, Yu and
                   Ma, Yubo and Yang, Baosong and Wei, Haoran and Tang, Jialong
                   and Lin, Huan and Gao, Ruize and Li, Tianhao and Cao, Qian
                   and Ren, Xuancheng and Deng, Xiaodong and Yang, An and
                   Huang, Fei and Liu, Dayiheng and Zhou, Jingren",
  month         =  may,
  year          =  2026,
  copyright     = "http://arxiv.org/licenses/nonexclusive-distrib/1.0/",
  archivePrefix = "arXiv",
  primaryClass  = "cs.CL",
  eprint        = "2605.11887",
  journal       = "arXiv preprint",
  url           = "https://arxiv.org/pdf/2605.11887v1"
}

@ARTICLE{He2024-vk,
  title         = "Llama Scope: Extracting millions of features from
                   {Llama-3.1-8B} with sparse Autoencoders",
  author        = "He, Zhengfu and Shu, Wentao and Ge, Xuyang and Chen, Lingjie
                   and Wang, Junxuan and Zhou, Yunhua and Liu, Frances and Guo,
                   Qipeng and Huang, Xuanjing and Wu, Zuxuan and Jiang, Yu-Gang
                   and Qiu, Xipeng",
  month         =  oct,
  year          =  2024,
  copyright     = "http://creativecommons.org/licenses/by/4.0/",
  archivePrefix = "arXiv",
  primaryClass  = "cs.LG",
  eprint        = "2410.20526",
  journal       = "arXiv preprint",
  url           = "https://arxiv.org/abs/2410.20526"
}

@misc{Qwennext25,
  title  = {{Qwen3-Next: Towards Ultimate Training \& Inference Efficiency}},
  author = {{Qwen Team}},
  year   = {2025},
  url    = {https://qwen.ai/blog?id=e34c4305036ce60d55a0791b170337c2b70ae51d&from=home.latest-research-list}
}

@ARTICLE{Heimersheim2024-gs,
  title         = "How to use and interpret activation patching",
  author        = "Heimersheim, Stefan and Nanda, Neel",
  month         =  apr,
  year          =  2024,
  copyright     = "http://arxiv.org/licenses/nonexclusive-distrib/1.0/",
  archivePrefix = "arXiv",
  primaryClass  = "cs.LG",
  eprint        = "2404.15255",
  journal       = "arXiv preprint",
  url           = "https://arxiv.org/abs/2404.15255"
}

@inbook{Biaozhang1029,
author = {Zhang, Biao and Sennrich, Rico},
title = {Root mean square layer normalization},
year = {2019},
publisher = {Curran Associates Inc.},
address = {Red Hook, NY, USA},
booktitle = {Proceedings of the 33rd International Conference on Neural Information Processing Systems},
articleno = {1110},
numpages = {12},
url = {https://dl.acm.org/doi/abs/10.5555/3454287.3455397}
}

@ARTICLE{Touvron2023-ww,
  title         = "{LLaMA}: Open and efficient foundation language models",
  author        = "Touvron, Hugo and Lavril, Thibaut and Izacard, Gautier and
                   Martinet, Xavier and Lachaux, Marie-Anne and Lacroix,
                   Timoth{\'e}e and Rozi{\`e}re, Baptiste and Goyal, Naman and
                   Hambro, Eric and Azhar, Faisal and Rodriguez, Aurelien and
                   Joulin, Armand and Grave, Edouard and Lample, Guillaume",
  month         =  feb,
  year          =  2023,
  copyright     = "http://creativecommons.org/licenses/by/4.0/",
  archivePrefix = "arXiv",
  primaryClass  = "cs.CL",
  eprint        = "2302.13971",
  journal       = "arXiv preprint",
  url           = "https://arxiv.org/abs/2302.13971"
}

%\bibliography{custom}

\appendix
\counterwithin{figure}{section}
\setcounter{figure}{0} % Reset figure counter
\counterwithin{table}{section}
\setcounter{table}{0} % Reset figure counter

\section{Appendix: Contrastive Pairs for Gemma} \label{sec:pairs}

\begin{table}[h]
\centering
\resizebox{0.65\linewidth}{!}{%
\begin{tabular}{llr}
\toprule
\textbf{LOW} & \textbf{HIGH} & \textbf{Count} \\
\midrule
nob\_Latn & eng\_Latn & 140 \\
nob\_Latn & zho\_Hans & 125 \\
hin\_Deva & zho\_Hans & 133 \\
hin\_Deva & eng\_Latn & 145 \\
npi\_Deva & hin\_Deva & 61 \\
ukr\_Cyrl & rus\_Cyrl & 60 \\
nob\_Latn & swe\_Latn & 112 \\
khm\_Khmr & eng\_Latn & 88 \\
khm\_Khmr & zho\_Hans & 79 \\
\bottomrule
\end{tabular}%
}
\caption{Contrastive LOW/HIGH languages for \texttt{gemma-3-4b-pt} on Belebele. \textbf{Count} is the number of documents answered incorrectly in the LOW language and correctly in the HIGH language. nob\_Latn and npi\_Deva are closely related languages to nno\_Latn and kas\_Deva, which are not available in Belebele.}
\label{tab:contrasting-MC}
\end{table}

\begin{table}[h]
\centering
\resizebox{0.99\linewidth}{!}{%
\begin{tabular}{llcc}
\toprule
\textbf{LOW} & \textbf{HIGH} & \textbf{Count} & \textbf{$\tau_{\text{lo}}$--$\tau_{\text{hi}}$} \\
\midrule
\multicolumn{4}{l}{\textit{Same target (different source)}} \\
\midrule
cmn\_Hans-nno\_Latn & eng\_Latn-nno\_Latn & 262 & 0.60--0.80 \\
kas\_Deva-eng\_Latn & hin\_Deva-eng\_Latn & 550 & 0.50--0.90  \\
khm\_Khmr-eng\_Latn & cmn\_Hans-eng\_Latn & 99 & 0.60--0.80 \\
ukr\_Cyrl-eng\_Latn & rus\_Cyrl-eng\_Latn & 85 & 0.85--0.95 \\
nno\_Latn-eng\_Latn & swe\_Latn-eng\_Latn & 60 & 0.90--1.00 \\
kas\_Deva-cmn\_Hans & ukr\_Cyrl-cmn\_Hans & 188 & 0.30--0.90 \\
khm\_Khmr-deu\_Latn & rus\_Cyrl-deu\_Latn & 159 & 0.60--0.80 \\
\midrule
\multicolumn{4}{l}{\textit{Different target, same source}} \\
\midrule
cmn\_Hans-nno\_Latn & cmn\_Hans-eng\_Latn & 68 & 0.60--0.80 \\
eng\_Latn-kas\_Deva & eng\_Latn-hin\_Deva & 269 & 0.50--0.80 \\
eng\_Latn-khm\_Khmr & eng\_Latn-cmn\_Hans & 145 & 0.40--0.85 \\
eng\_Latn-ukr\_Cyrl & eng\_Latn-rus\_Cyrl & 151 & 0.80--0.95 \\
eng\_Latn-nno\_Latn & eng\_Latn-swe\_Latn & 148 & 0.80--0.95 \\
cmn\_Hans-kas\_Deva & cmn\_Hans-ukr\_Cyrl & 309 & 0.20--0.95 \\
deu\_Latn-khm\_Khmr & deu\_Latn-rus\_Cyrl & 177 & 0.30--0.90 \\
eng\_Latn-kas\_Deva & eng\_Latn-rus\_Cyrl & 135 & 0.15--0.90 \\
eng\_Latn-khm\_Khmr & eng\_Latn-hin\_Deva & 98 & 0.60--0.80 \\
\midrule
\multicolumn{4}{l}{\textit{Different target, disjoint}} \\
\midrule
kas\_Deva-cmn\_Hans & hin\_Deva-eng\_Latn & 500 & 0.40--0.90 \\
ukr\_Cyrl-cmn\_Hans & rus\_Cyrl-eng\_Latn & 125 & 0.60--0.80 \\
nno\_Latn-rus\_Cyrl & cmn\_Hans-eng\_Latn & 65 & 0.75--0.90 \\
nno\_Latn-rus\_Cyrl & deu\_Latn-eng\_Latn & 93 & 0.70--0.90 \\
nno\_Latn-cmn\_Hans & swe\_Latn-eng\_Latn & 129 & 0.60--0.80 \\
kas\_Deva-cmn\_Hans & rus\_Cyrl-deu\_Latn & 200 & 0.20--0.95 \\
\bottomrule
\end{tabular}%
}
\caption{Contrastive LOW/HIGH translation directions for \texttt{gemma-3-4b-pt}, grouped by the relation between the two directions. \textbf{Count} is the number of segments surviving the selection rule of Eq.~\ref{eq:selection} under the per-pair xCOMET-XL thresholds $\tau_{\text{lo}}$ (LOW maximum) and $\tau_{\text{hi}}$ (HIGH minimum).}
\label{tab:contrasting-MT}
\end{table}

\section{Appendix: LLM Inference Details} \label{sec:inf_details}

\begin{figure*}[t]
  \centering
  \includegraphics[width=0.325\textwidth]{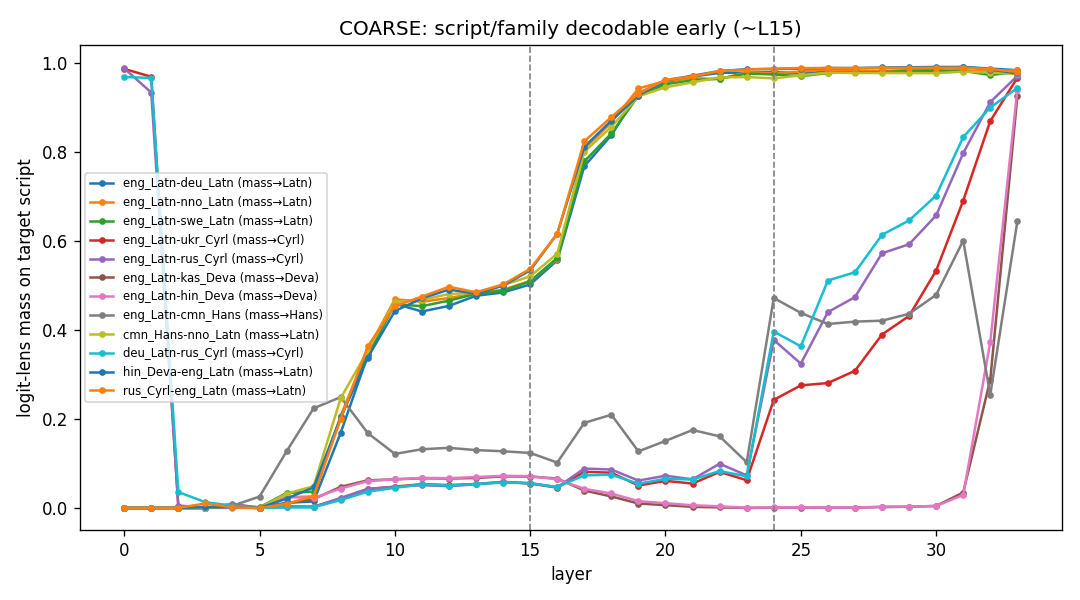} \hfill
  \includegraphics[width=0.325\textwidth]{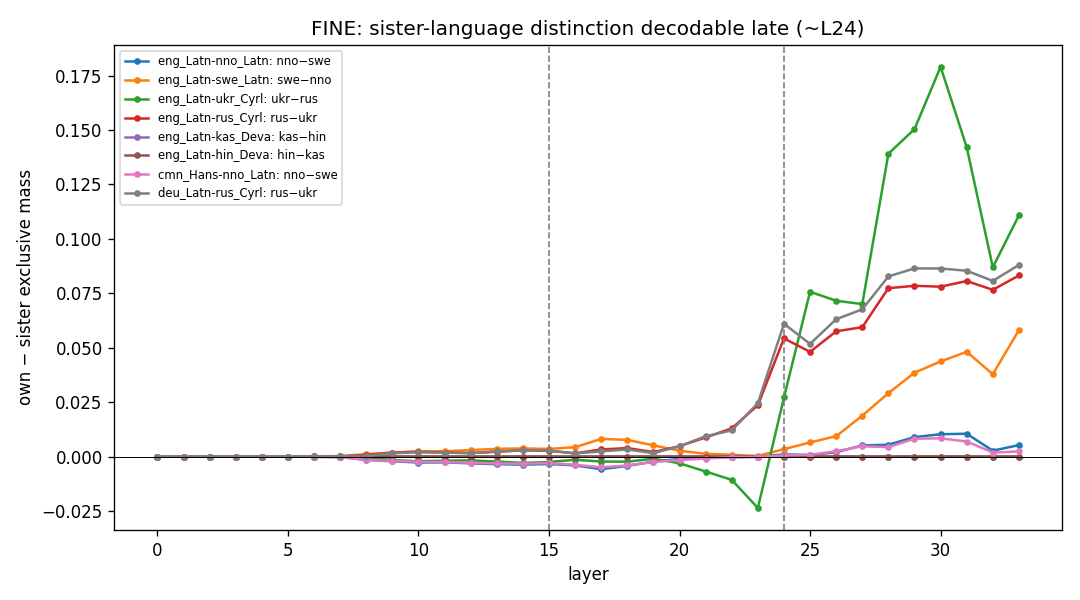} \hfill
  \includegraphics[width=0.325\textwidth]{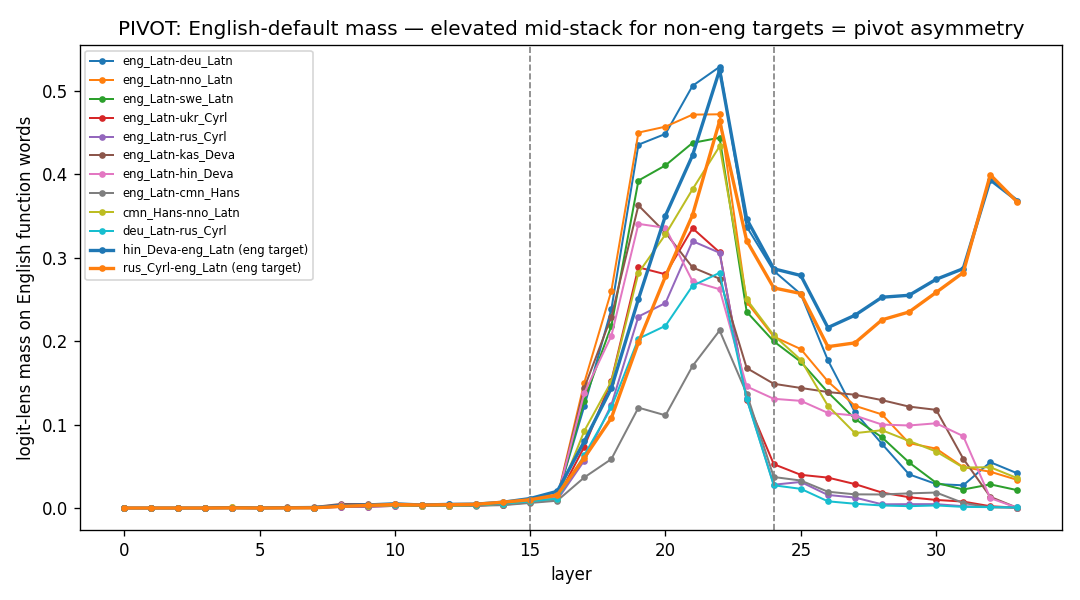}
  \caption{Logit lens decodes the coarse script (left), fine sister family (middle) and the English pivot (right).}
  \label{fig.logit_lens}
\end{figure*}

The MT task is performed under 5-shot setting, and the five demonstrations are selected from the FLORES+ \texttt{dev} set. The demonstrations are fixed for each direction, and their identifiers are stored with the outputs so that every later stage rebuilds byte-identical prompts. We use greedy decoding with a maximum generation length of $512$ tokens, truncated at the first newline. Translations are generated with vLLM~\citep{kwon2023vllm}. Initial accuracies come from lm-evaluation-harness~\citep{eval-harness}, and our intervention code re-implements the identical template and scoring rule.

All experiments ran on one NVIDIA GH200 per job. Under vLLM, one translation direction ($1{,}012$ FLORES+ \texttt{devtest} segments, $5$-shot, greedy, newline-truncated) takes $8$\,s on average for \texttt{gemma-3-4b-pt} and \texttt{Llama-3.1-8B} and $38$\,s for \texttt{Qwen3.5-9B-Base}. Summing over $32$ FLORES+ directions plus the NTREX-128 top-up, generation costs roughly $0.6$ GPU-hours across all three models. For Belebele, lm-evaluation-harness scores one language in about $50$\,s, so the full sweep is $8$--$9$ minutes per model. The whole-pipeline cost is mainly set by the patching sweeps in Appendix~\ref{sec:patch_procedure}.

\section{Appendix: Patching Procedure} \label{sec:patch_procedure}

The patching sweep runs in two passes over the instance set per pair. The first pass is intervention-free and builds the \emph{donor bank}. For every instance, it caches $\mathbf{h}_i^{(\ell)}$ at every layer from a forward pass on $x_i^{\textsc{high}}$ into the bank. When the additive condition is requested, $\tilde{\mathbf{h}}_i^{(\ell)}$ from a forward pass on $x_i^{\textsc{low}}$ is also cached, since the steering direction needs the LOW mean. The donor assignment $\pi$ and the layer-wise means are computed from these caches over the \emph{full} instance list, so they do not depend on how the second pass is distributed across jobs.

The second pass establishes the clean baselines and runs the patched sweep. For each instance, it first records the unpatched teacher-forced negative log-likelihood of the LOW reference and one unpatched greedy generation, then repeats both, under every donor condition and every layer. In the generation runs, the intervention is applied \emph{only at the first decoding step}. All later steps run unmodified, so any persistent change in the output is a downstream consequence of a single-position, single-layer write propagated through the KV cache.

The interventions use nnsight~\citep{Fiotto-Kaufman2024-dj} in \texttt{bfloat16} with eager attention (which might lead to numerically different results from lm-evaluation-harness or vLLM in Figure~\ref{fig:decomp}). For the four-donor MT sweep of \S\ref{sec:patching}, the median pair costs $25$ GPU-hours on \texttt{gemma-3-4b-pt} ($18$ layers), $75$ on \texttt{Qwen3.5-9B-Base} ($20$ layers) and $207$ on \texttt{Llama-3.1-8B} ($20$ layers). Belebele costs $0.3$--$0.8$ GPU-hours per contrastive pair. Patching the SAE-derived vectors costs a median of $7$ GPU-hours per pair on Gemma, $46$ on Qwen and $180$ on Llama. The sparse flip conditions, run at two or three layers, cost $3$ GPU-hours per pair. Summing over all pairs, tasks and models, the interventions reported in this paper cost a total of approximately $16{,}500$ GPU hours. 

\section{Appendix: Logit Lens}
\label{sec:lens}

We apply the final RMSNorm~\citep{Biaozhang1029} and the unembedding to each layer's last-token residual and sum the next-token probability over three token masks: 1) Unicode blocks of language scripts (coarse), 2) characters exclusive to one sister language (fine), and 3) English-only function words (pivot). We use the same prompts as those in the patching sweeps when running the probes, so the decodability and patching onsets are measured on the same objects. Figure~\ref{fig.logit_lens} shows the results for the coarse, fine and pivot categories.

For \texttt{eng}$\rightarrow$\texttt{nno}, Latin-script mass is flat near $0.50$ through L15 and then steps to $0.56$ (L16), $0.81$ (L17) and $0.94$ (L19), while English function-word mass rises from $0.012$ at L16 to a peak of $0.47$ at L22 and then \emph{decays} to $0.03$ by L33. Non-Latin targets show the complement: Cyrillic mass for \texttt{eng}$\rightarrow$\texttt{rus} stays at $0.05$--$0.10$ through L23 before reaching $0.38$ at L24 and $0.97$ at L33; Han mass for \texttt{eng}$\rightarrow$\texttt{cmn} rises similarly at L24; and sister-exclusive masses separate only from L24. This mostly reproduces the patching onset map of $\S$\ref{non_eng_onset}: Latin and English rise at L16, whereas Cyrillic, Han and sister-exclusive masses stay flat until L24. Both the logit lens and patching therefore indicate that the mid-layer onset is the assertion of an English and high-resource Latin default rather than a generic target commitment.

\section{Appendix: Experimental Setup for Qwen and Llama} \label{sec:qwen_llama}

Everything not listed here is identical to the \texttt{gemma-3-4b-pt} setup of \S\ref{sec:setup}. Only the model, the SAEs, the swept layers and the per-pair thresholds for Llama differ. 

\paragraph{Models and SAEs.} \texttt{Qwen3.5-9B-Base}~\citep{qwen3.5} has $32$ layers, interleaved three linear-attention layers per full-attention layer, with a $4096$-dimensional residual stream. Its SAEs are \texttt{qwen-scope-3.5-9b-base-w64k-l50} (\texttt{sae\_id} \texttt{layer\{L\}}), TopK with $k{=}50$ and width $65{,}536$. \texttt{Llama-3.1-8B}~\citep{Grattafiori2024-ot} has $32$ layers with a $4096$-dimensional residual stream and uniform full attention. Its SAEs are \texttt{llama\_scope\_lxr\_32x} (\texttt{sae\_id} \texttt{l\{L\}r\_32x}), TopK $k{=}50$ and width $131{,}072$. The $32\times$ expansion is chosen as the closest analogue of Gemma Scope 2's $262$k. In the feature analyses, we rank the top $30$ features per layer for these two models against $50$ for Gemma, and use a shared-activation threshold of $0.7$ for Qwen against $0.9$ for Gemma and Llama, since the TopK dictionaries put fewer features above threshold on any single vector.

\paragraph{Thresholds.} The per-pair xCOMET-XL thresholds for Qwen are \emph{identical} to Gemma's (Table~\ref{tab:contrasting-MT}), so we omit that column from Table~\ref{tab:qwen-MT}. Since xCOMET-XL scores are distributed differently for Llama, we calibrate its thresholds in Table~\ref{tab:llama-MT}. This admits more instances per pair, resulting in the largest counts among the three. The number of instances of each contrastive pair on Belebele for both models are listed in Table~\ref{tab:qwen-llama-MC}. 

\paragraph{SAE analyses.} \label{sec:qwen_sae} Re-running the sparse conditions at L18 for Qwen, where the flip happens, puts the dense donors at $0.50$--$0.79$ of outputs in English on all six English-HIGH pairs, but the sparse interventions do not follow: barely zero percent of the output becomes English for all conditions. The shared/instance-specific analysis does not replicate either, since Qwen-Scope's TopK SAEs ($k{=}50$ out of $64$k) reconstruct only $45$--$73\%$ of the residual at the layers of interest, against $90$--$96\%$ for Gemma Scope 2. The \textit{denoised} vector is therefore dominated by reconstruction error rather than by the shared/instance-specific split, as we observe \textit{matched\_denoised} fails to preserve the flip ($0.29$ vs.\ $0.63$ for \textit{matched} at L18) at a level barely distinguishable from the \textit{random\_ablate} control ($0.29$).

Llama-Scope's TopK SAEs reconstruct only $26-55\%$ of the last-token residual at this position across the swept layers, a range sitting far below Qwen and Gemma's. No feature is identified as exclusively language-typed on the max-activating-feature dashboard at that fidelity. The \textit{denoised} vector is dominated by reconstruction error. We therefore do not report the sparse-flip test and the \textit{denoised} sweep for Llama at all.

We interpret this as a limit on the instrument, not a contradiction of our results on Gemma. The three models make reconstruction fidelity the variable that governs whether a feature-level decomposition means anything: $0.90$--$0.96$ on Gemma gives an interpretable split; $0.45$--$0.73$ on Qwen gives one dominated by reconstruction errors; and $0.26$--$0.55$ on Llama gives none at all. However, the patching-level account, which needs no autoencoder, transfers across all three models regardless.

\begin{table}[h]
\centering
\resizebox{0.86\linewidth}{!}{%
\begin{tabular}{llc}
\toprule
\textbf{LOW} & \textbf{HIGH} & \textbf{Count} \\
\midrule
\multicolumn{3}{l}{\textit{Same target (different source)}} \\
\midrule
cmn\_Hans-nno\_Latn & eng\_Latn-nno\_Latn & 174 \\
kas\_Deva-cmn\_Hans & ukr\_Cyrl-cmn\_Hans & 247 \\
kas\_Deva-eng\_Latn & hin\_Deva-eng\_Latn & 498 \\
khm\_Khmr-deu\_Latn & rus\_Cyrl-deu\_Latn & 227 \\
khm\_Khmr-eng\_Latn & cmn\_Hans-eng\_Latn & 201 \\
nno\_Latn-eng\_Latn & swe\_Latn-eng\_Latn & 28 \\
ukr\_Cyrl-eng\_Latn & rus\_Cyrl-eng\_Latn & 55 \\
\midrule
\multicolumn{3}{l}{\textit{Different target, same source}} \\
\midrule
cmn\_Hans-kas\_Deva & cmn\_Hans-ukr\_Cyrl & 276 \\
cmn\_Hans-nno\_Latn & cmn\_Hans-eng\_Latn & 202 \\
deu\_Latn-khm\_Khmr & deu\_Latn-rus\_Cyrl & 327 \\
eng\_Latn-kas\_Deva & eng\_Latn-hin\_Deva & 140 \\
eng\_Latn-kas\_Deva & eng\_Latn-rus\_Cyrl & 206 \\
eng\_Latn-khm\_Khmr & eng\_Latn-cmn\_Hans & 336 \\
eng\_Latn-khm\_Khmr & eng\_Latn-hin\_Deva & 55 \\
eng\_Latn-nno\_Latn & eng\_Latn-swe\_Latn & 81 \\
eng\_Latn-ukr\_Cyrl & eng\_Latn-rus\_Cyrl & 112 \\
\midrule
\multicolumn{3}{l}{\textit{Different target, disjoint}} \\
\midrule
kas\_Deva-cmn\_Hans & hin\_Deva-eng\_Latn & 457 \\
kas\_Deva-cmn\_Hans & rus\_Cyrl-deu\_Latn & 154 \\
nno\_Latn-cmn\_Hans & swe\_Latn-eng\_Latn & 62 \\
nno\_Latn-rus\_Cyrl & cmn\_Hans-eng\_Latn & 157 \\
nno\_Latn-rus\_Cyrl & deu\_Latn-eng\_Latn & 68 \\
ukr\_Cyrl-cmn\_Hans & rus\_Cyrl-eng\_Latn & 64 \\
\bottomrule
\end{tabular}%
}
\caption{Contrastive LOW/HIGH translation directions for \texttt{Qwen3.5-9B-Base} ($22$ pairs). Thresholds are identical to Gemma's in Table~\ref{tab:contrasting-MT}.}
\label{tab:qwen-MT}
\end{table}

\begin{table}[h]
\centering
\resizebox{0.99\linewidth}{!}{%
\begin{tabular}{llcc}
\toprule
\textbf{LOW} & \textbf{HIGH} & \textbf{Count} & \textbf{$\tau_{\text{lo}}$--$\tau_{\text{hi}}$} \\
\midrule
\multicolumn{4}{l}{\textit{Same target (different source)}} \\
\midrule
cmn\_Hans-nno\_Latn & eng\_Latn-nno\_Latn & 65 & 0.70--0.90 \\
kas\_Deva-cmn\_Hans & ukr\_Cyrl-cmn\_Hans & 552 & 0.40--0.70 \\
kas\_Deva-eng\_Latn & hin\_Deva-eng\_Latn & 793 & 0.60--0.80 \\
khm\_Khmr-deu\_Latn & rus\_Cyrl-deu\_Latn & 280 & 0.70--0.90 \\
khm\_Khmr-eng\_Latn & cmn\_Hans-eng\_Latn & 262 & 0.70--0.90 \\
nno\_Latn-eng\_Latn & swe\_Latn-eng\_Latn & 42 & 0.87--0.97 \\
ukr\_Cyrl-eng\_Latn & rus\_Cyrl-eng\_Latn & 23 & 0.87--0.97 \\
\midrule
\multicolumn{4}{l}{\textit{Different target, same source}} \\
\midrule
cmn\_Hans-kas\_Deva & cmn\_Hans-ukr\_Cyrl & 787 & 0.40--0.70 \\
cmn\_Hans-nno\_Latn & cmn\_Hans-eng\_Latn & 109 & 0.75--0.95 \\
deu\_Latn-khm\_Khmr & deu\_Latn-rus\_Cyrl & 879 & 0.50--0.75 \\
eng\_Latn-kas\_Deva & eng\_Latn-hin\_Deva & 491 & 0.30--0.60 \\
eng\_Latn-kas\_Deva & eng\_Latn-rus\_Cyrl & 907 & 0.50--0.75 \\
eng\_Latn-khm\_Khmr & eng\_Latn-cmn\_Hans & 765 & 0.30--0.60 \\
eng\_Latn-khm\_Khmr & eng\_Latn-hin\_Deva & 433 & 0.30--0.60 \\
eng\_Latn-nno\_Latn & eng\_Latn-swe\_Latn & 46 & 0.75--0.95 \\
eng\_Latn-ukr\_Cyrl & eng\_Latn-rus\_Cyrl & 45 & 0.75--0.95 \\
\midrule
\multicolumn{4}{l}{\textit{Different target, disjoint}} \\
\midrule
kas\_Deva-cmn\_Hans & hin\_Deva-eng\_Latn & 857 & 0.60--0.80 \\
kas\_Deva-cmn\_Hans & rus\_Cyrl-deu\_Latn & 929 & 0.60--0.80 \\
nno\_Latn-cmn\_Hans & swe\_Latn-eng\_Latn & 282 & 0.75--0.95 \\
nno\_Latn-rus\_Cyrl & cmn\_Hans-eng\_Latn & 85 & 0.75--0.95 \\
nno\_Latn-rus\_Cyrl & deu\_Latn-eng\_Latn & 133 & 0.75--0.95 \\
ukr\_Cyrl-cmn\_Hans & rus\_Cyrl-eng\_Latn & 283 & 0.75--0.95 \\
\bottomrule
\end{tabular}%
}
\caption{Contrastive LOW/HIGH translation directions for \texttt{Llama-3.1-8B} ($22$ pairs), with the re-derived per-pair xCOMET-XL thresholds.}
\label{tab:llama-MT}
\end{table}

\begin{table}[h]
\centering
\resizebox{0.78\linewidth}{!}{%
\begin{tabular}{llrr}
\toprule
\textbf{LOW} & \textbf{HIGH} & \textbf{Qwen} & \textbf{Llama} \\
\midrule
nob\_Latn & eng\_Latn & 98 & 241 \\
nob\_Latn & zho\_Hans & 89 & 218 \\
hin\_Deva & zho\_Hans & 165 & 157 \\
hin\_Deva & eng\_Latn & 177 & 180 \\
npi\_Deva & hin\_Deva & 74 & 79 \\
ukr\_Cyrl & rus\_Cyrl & 30 & 78 \\
nob\_Latn & swe\_Latn & 79 & 144 \\
khm\_Khmr & eng\_Latn & 74 & 116 \\
khm\_Khmr & zho\_Hans & 73 & 110 \\
\bottomrule
\end{tabular}%
}
\caption{Contrastive LOW/HIGH languages on Belebele for \texttt{Qwen3.5-9B-Base} and \texttt{Llama-3.1-8B}, over the same nine language contrasts used for Gemma (Table~\ref{tab:contrasting-MC}). \textbf{Count} is the number of documents answered incorrectly in the LOW language and correctly in the HIGH one. The selection criterion is binary and identical across models, so no thresholds apply.}
\label{tab:qwen-llama-MC}
\end{table}

\section{Appendix: Additional Figures and Tables} \label{sec:add_results}

\begin{figure*}[t]
  \centering
  \includegraphics[width=0.48\textwidth]{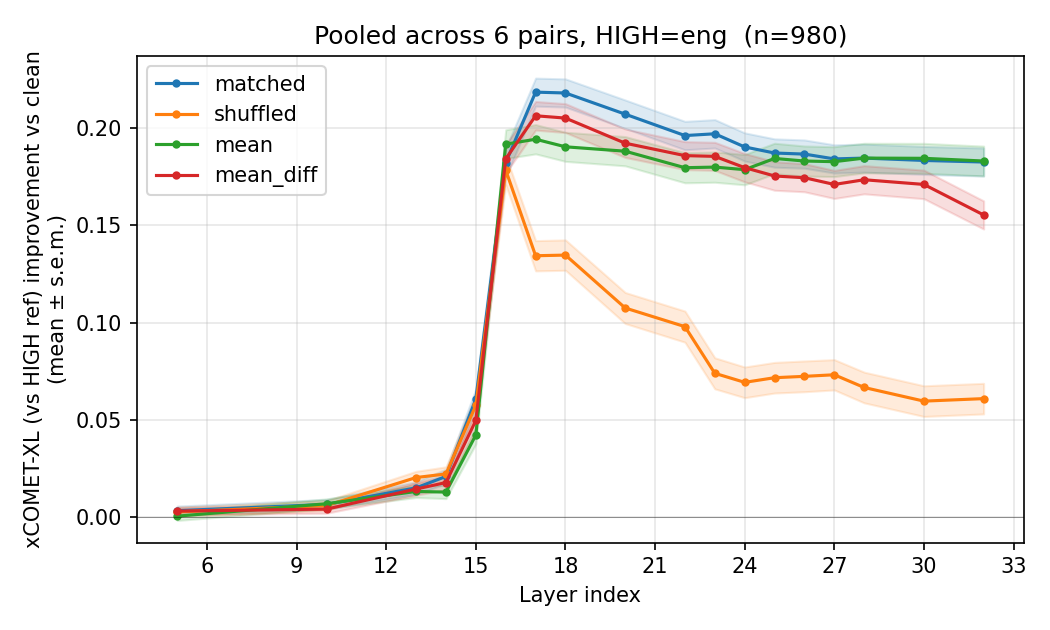}\hfill
  \includegraphics[width=0.48\textwidth]{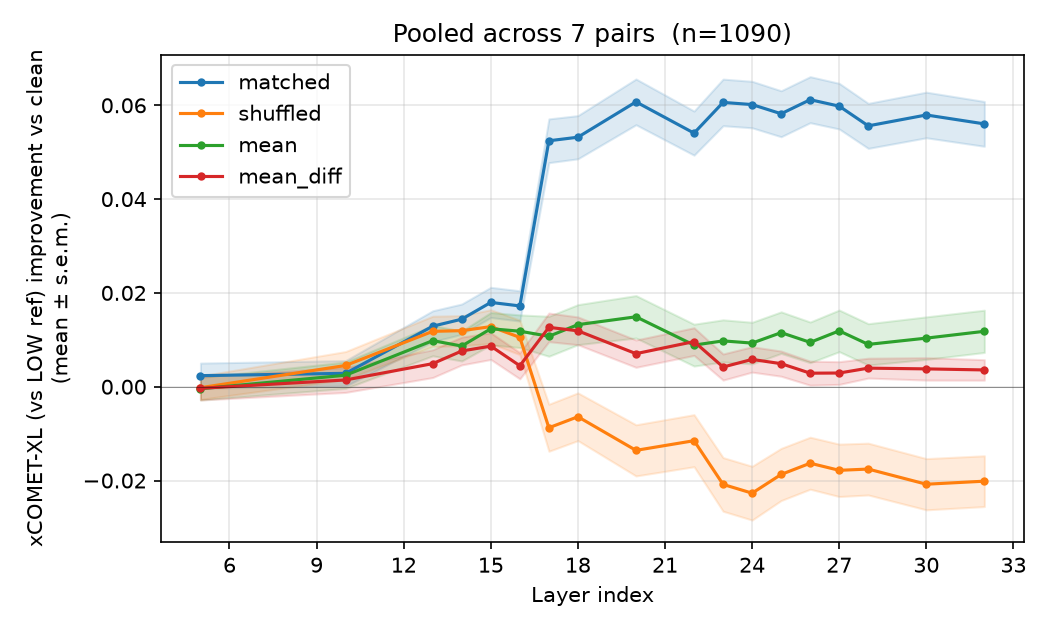}
  \caption{xCOMET-XL counterparts of Figure~\ref{fig:decomp}. \textbf{Left}: the six different-target pairs with English as HIGH target. \textbf{Right}: the seven same-target pairs. Both reproduce the onsets read off chrF++; the late-layer \textit{mean}-over-\textit{matched} ordering of the left panel of Figure~\ref{fig:decomp} does not reproduce here.}
  \label{fig.concat_xcomet_high_eng}
\end{figure*}

\begin{figure*}[h]
  \centering
  \includegraphics[width=0.325\textwidth]{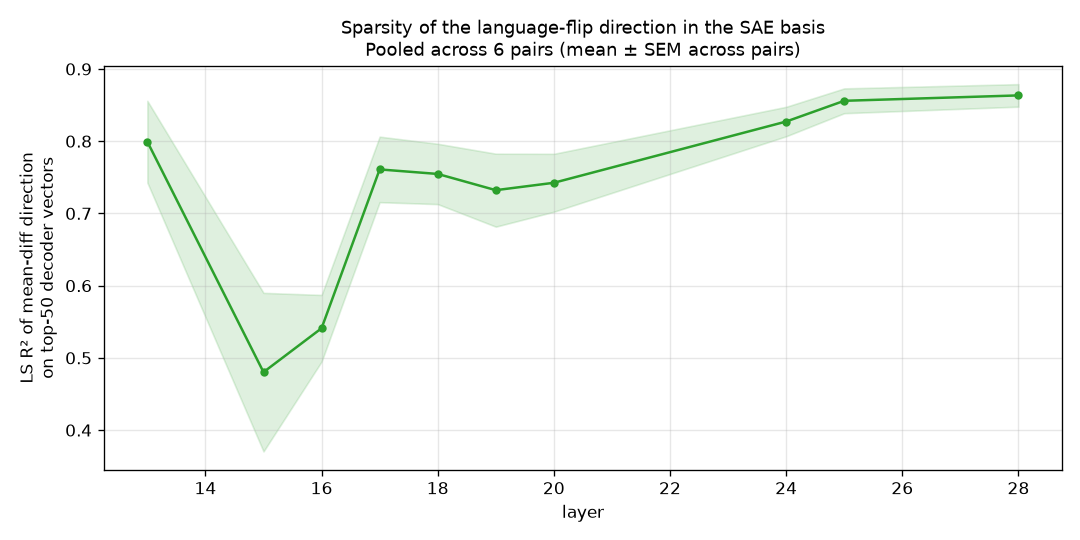}\hfill
  \includegraphics[width=0.325\textwidth]{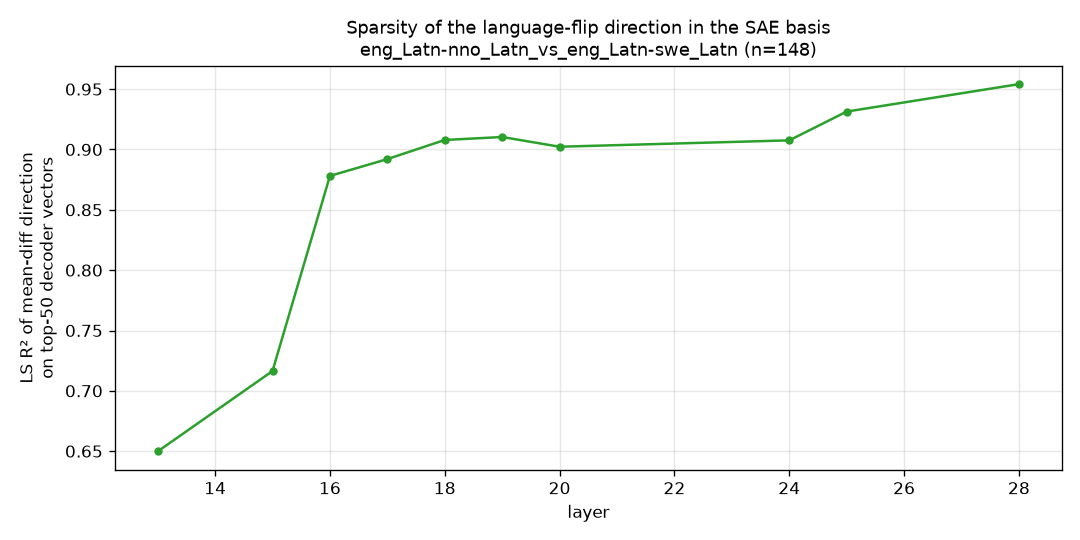}\hfill
  \includegraphics[width=0.325\textwidth]{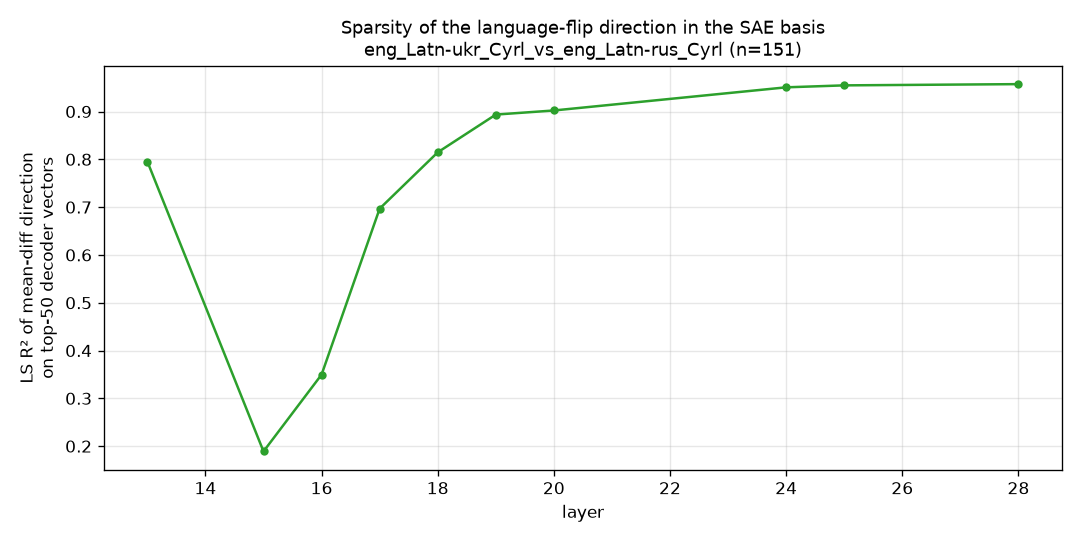}
  \caption{Cumulative least-squares $R^2$ of the residual \textit{mean\_diff} direction on its top-$50$ decoder vectors (Eq.~\ref{eq.rsquared}). \textbf{Left}: pooled across 6 pairs whose HIGH-target is English with the Standard Error of the Mean (SEM). The dip at L15--16 is where the English flip is causal. \textbf{Middle}: English-Nynorsk vs English-Swedish (n=148). \textbf{Right}: English-Ukrainian vs English-Russian (n=151).}
  \label{fig.sae_flip}
\end{figure*}

\begin{figure*}[h]
  \centering
  \includegraphics[width=0.9\textwidth]{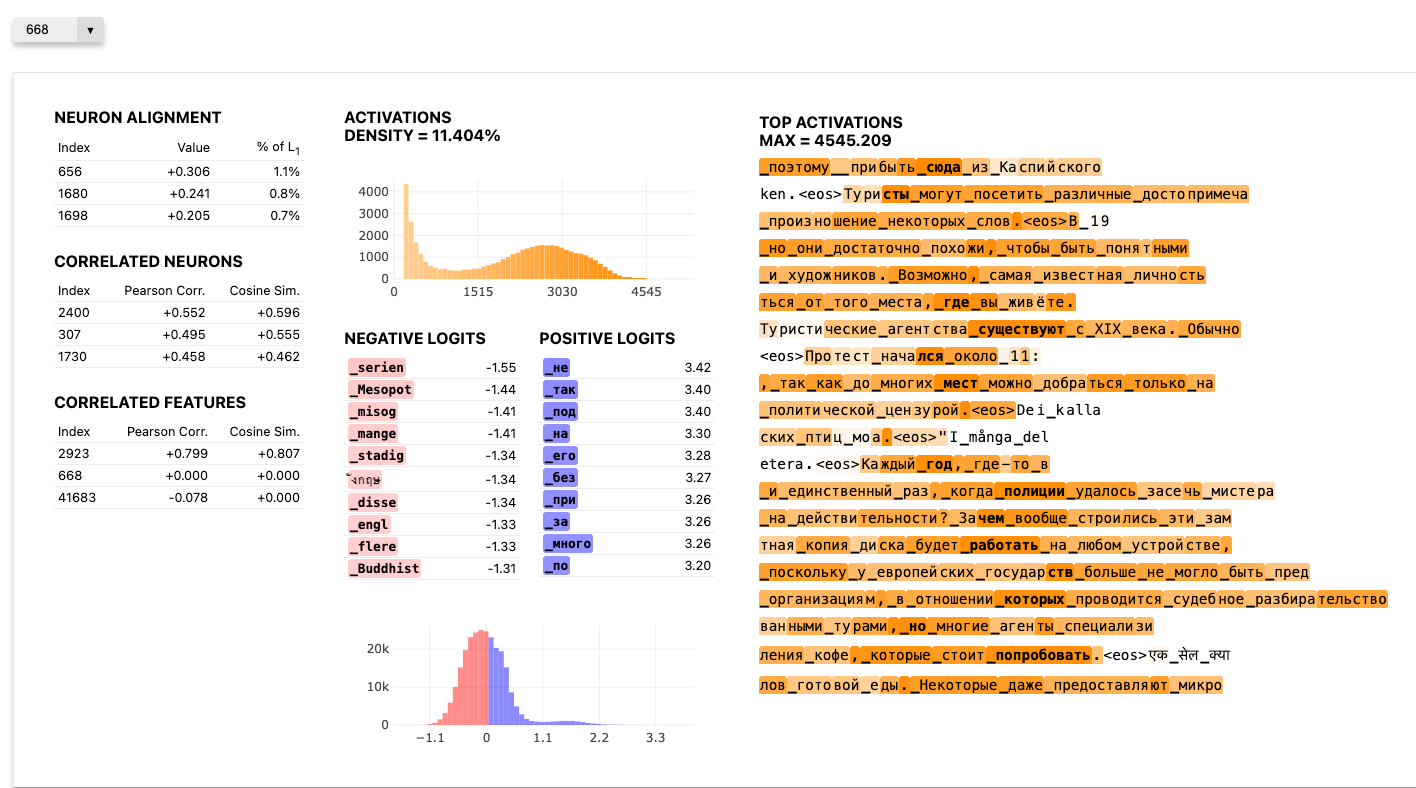}
  \caption{The dashboard to inspect active SAE feature IDs when generating tokens in a specific language.}
  \label{fig.dashboard}
\end{figure*}

\begin{figure*}[h]
  \centering
  \includegraphics[width=0.9\textwidth]{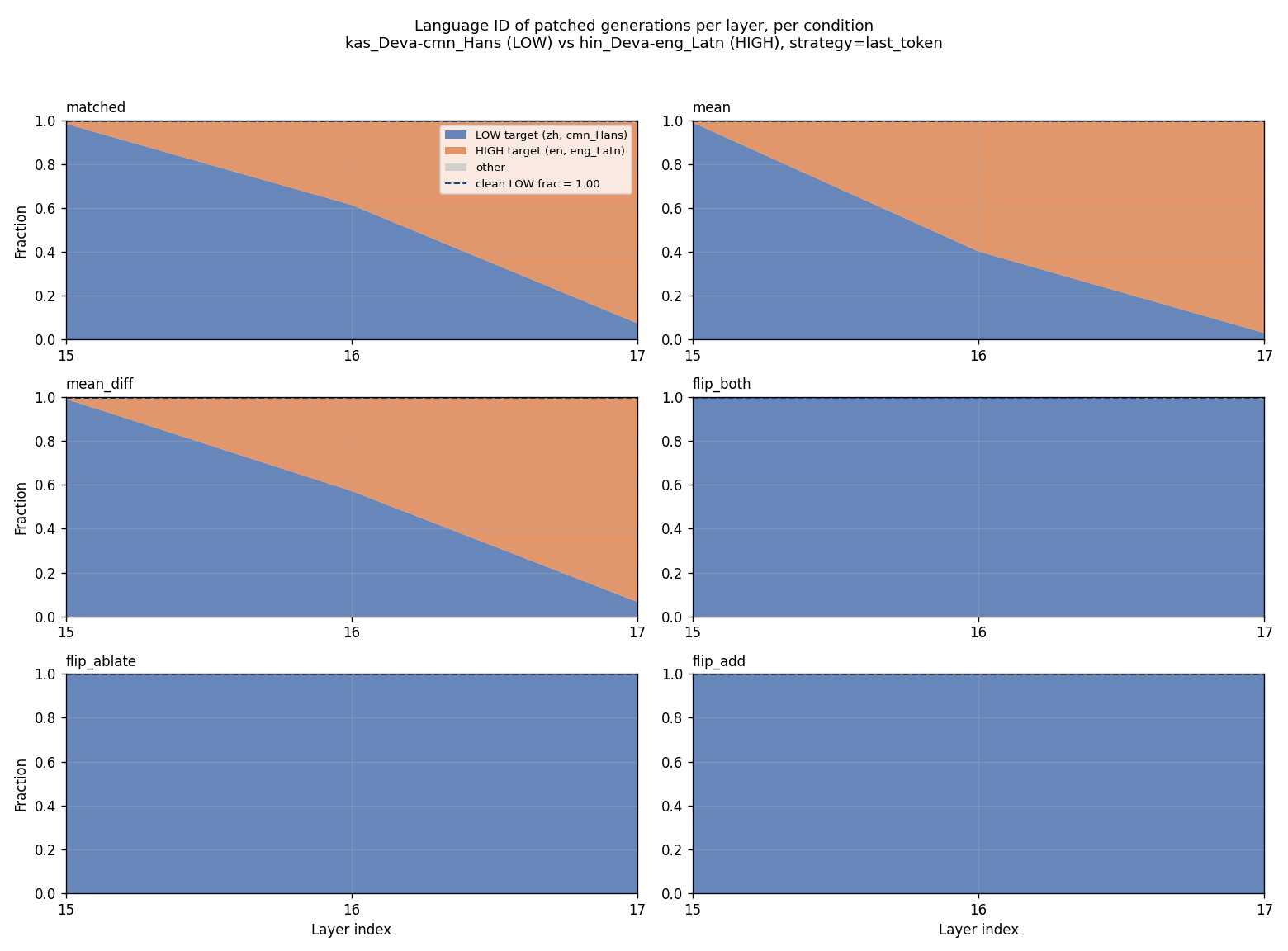}
  \caption{Patching by \textit{add} or \textit{ablate} the SAE feature IDs (or do both) to test English flip. \textit{fraction} reported by using fastText lid.176.}
  \label{fig.patching_flip_English}
\end{figure*}

\begin{figure*}[h]
  \centering
  \includegraphics[width=0.9\textwidth]{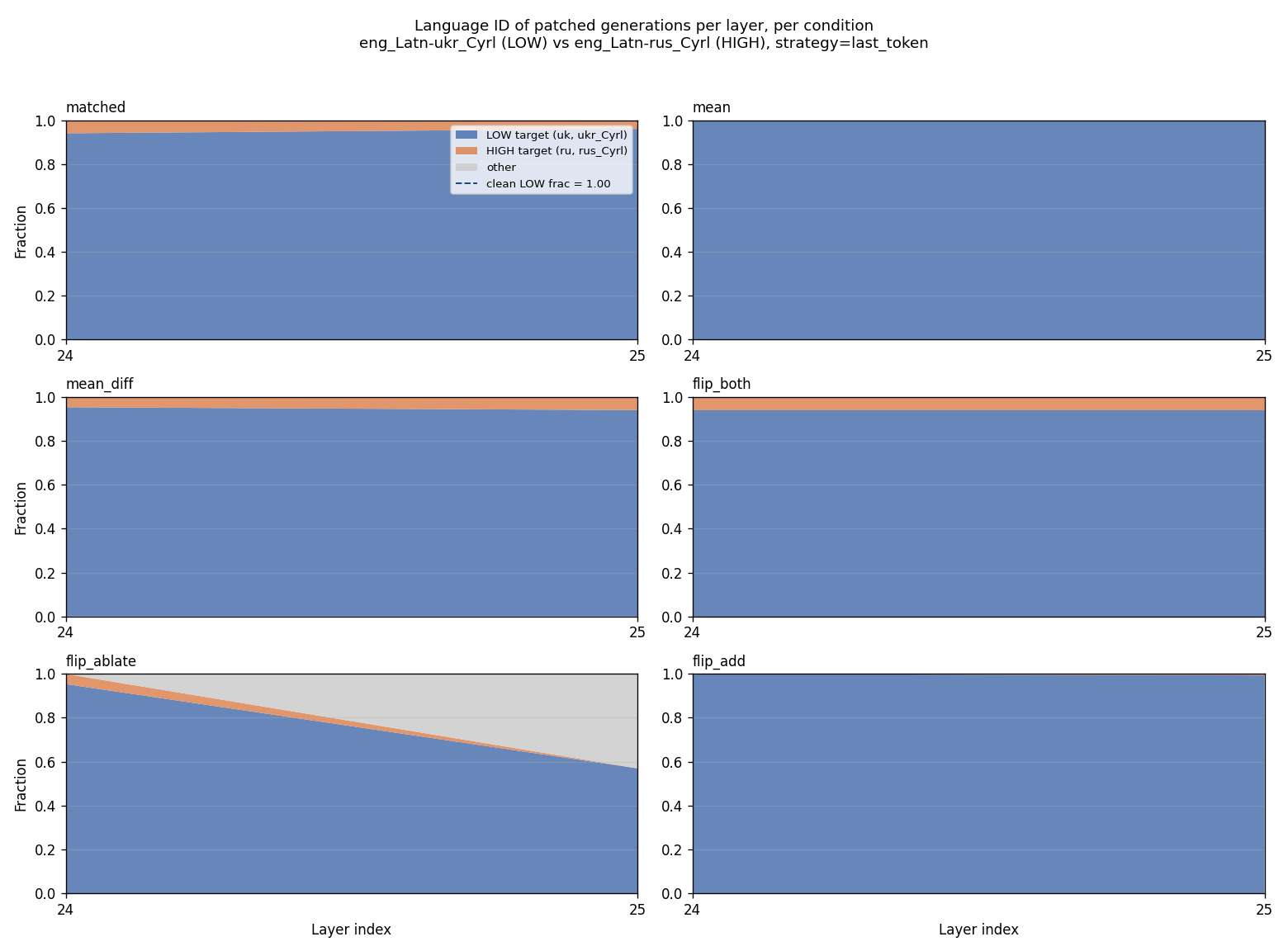}
  \caption{Patching by \textit{add} or \textit{ablate} the SAE feature IDs (or do both) to test sister language (\texttt{ukr}$\rightarrow$\texttt{rus}) flip. \textit{fraction} reported by using fastText lid.176.}
  \label{fig.patching_flip_sister}
\end{figure*}

\begin{figure*}[h]
  \centering
  \includegraphics[width=0.9\textwidth]{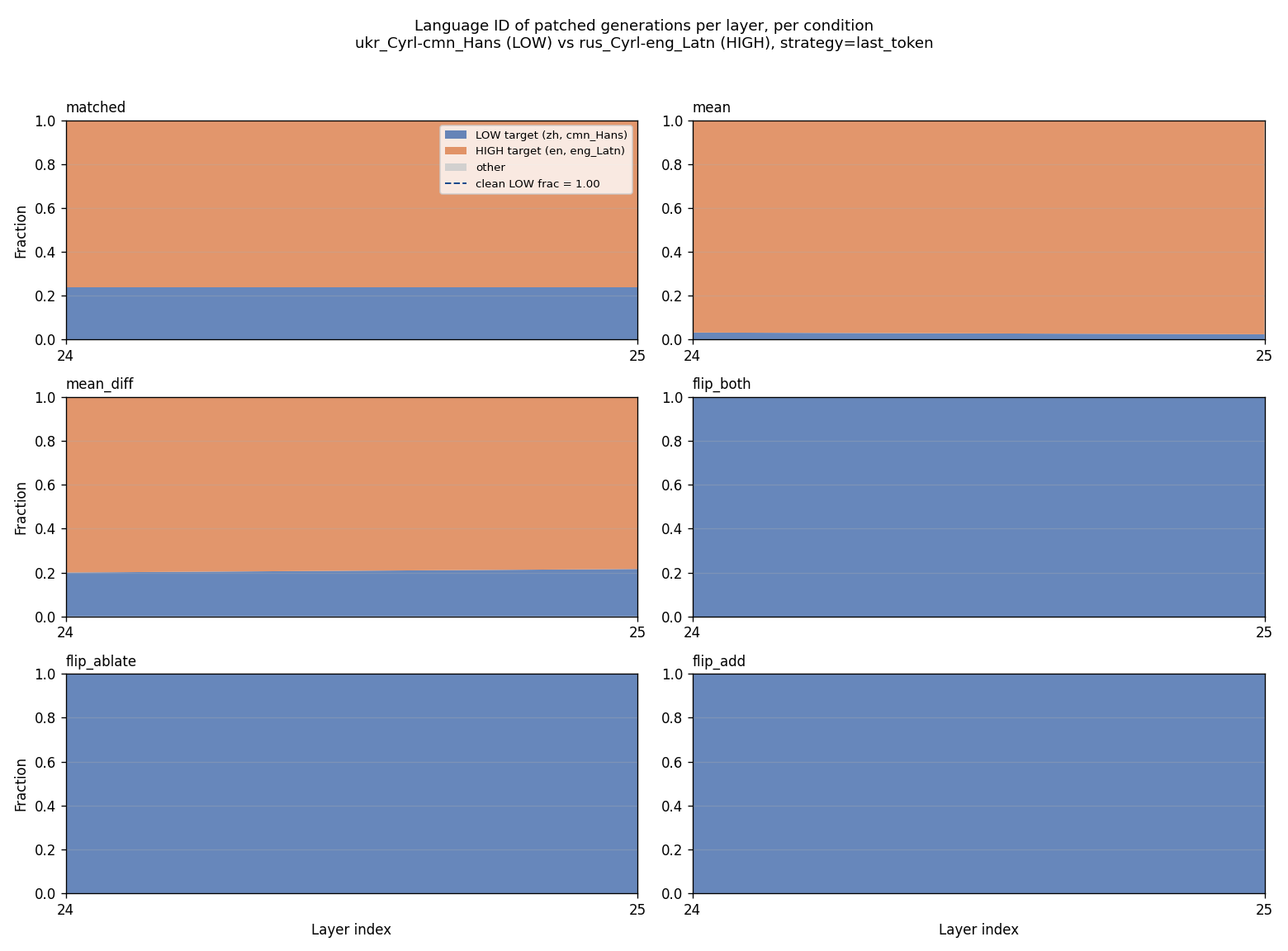}
  \caption{Patching by \textit{add} or \textit{ablate} the SAE feature IDs (or do both) for disjoint pairs with the source as sister languages (\texttt{ukr}$\rightarrow$\texttt{rus}). \textit{fraction} reported by using fastText lid.176.}
  \label{fig.patching_flip_sister_compare}
\end{figure*}

\begin{figure*}[t]
  \centering
  \includegraphics[width=1\columnwidth]{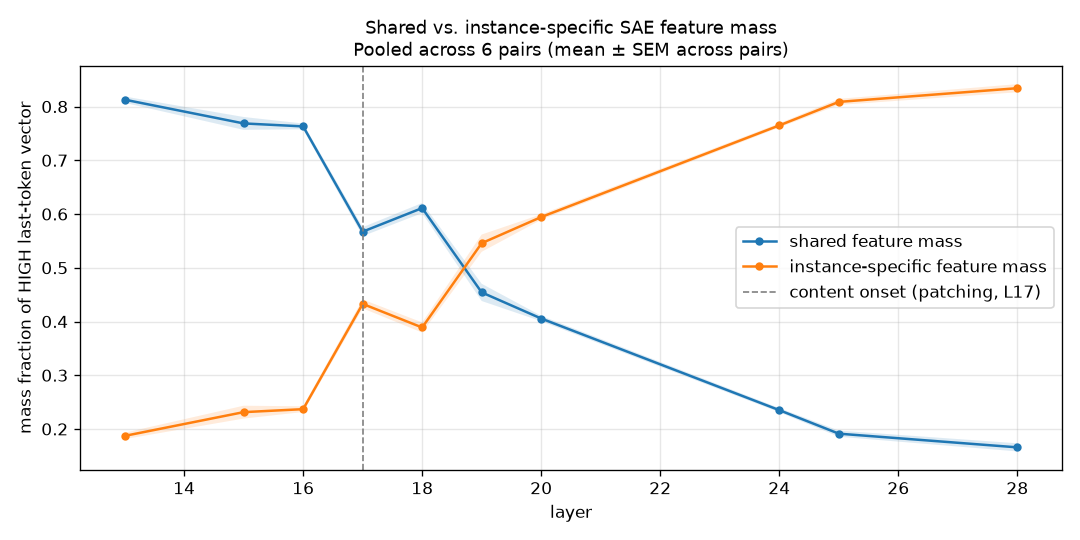}
  \caption{Shared vs.\ instance-specific SAE feature mass in the HIGH vectors. The crossover begins at the content-onset layer L17.}
  \label{fig:sae}
\end{figure*}

\begin{comment}
\begin{figure*}[t]
  \centering
  \includegraphics[width=0.45\textwidth]{pics/mt/eng_Latn-nno_Latn_vs_eng_Latn-swe_Latn/cross_ref_controls_curves_high.png} \hfill
  \includegraphics[width=0.45\textwidth]{pics/mt/eng_Latn-khm_Khmr_vs_eng_Latn-cmn_Hans/cross_ref_controls_curves_high.png}
  \caption{chrF++ scores against HIGH target reference when patching English-Swedish activations to English-Nynorsk (left) and English-Chinese activations to English-Khmer (right) at each layer of gemma-3-4b-pt.}
  \label{fig.close_distant}
\end{figure*}
\end{comment}

\end{document}